%% file: main.tex
\documentclass[conference]{IEEEtran}
\IEEEoverridecommandlockouts

\usepackage{cite}
\usepackage{amsmath,amssymb,amsfonts}
\usepackage{textcomp}
\usepackage{xcolor}
\def\BibTeX{{\rm B\kern-.05em{\sc i\kern-.025em b}\kern-.08em
    T\kern-.1667em\lower.7ex\hbox{E}\kern-.125emX}}

\usepackage[normalem]{ulem}
\usepackage[switch]{lineno}

\usepackage{tikz}

\usepackage{color}
\usepackage{listings}
\usepackage[]{hyperref}
\usepackage{amsmath,amsfonts}
\usepackage{mathtools}
\usepackage[ruled, vlined, linesnumbered]{algorithm2e}
\usepackage{circledsteps}
\usepackage{comment}
\usepackage{booktabs}
\usepackage{multirow}
\usepackage{graphicx}
\usepackage{makecell}
\usepackage{titlesec}
\usepackage{subcaption}
\usepackage{xcolor}
\usepackage{mathpartir}
\usepackage{adjustbox}
\usepackage{mdframed}
\usepackage{siunitx}
\usepackage{url}

\usepackage{pifont}
\usepackage[frozencache=true,cachedir=minted-cache]{minted} 
\usepackage{tikz}
\usetikzlibrary{tikzmark,fit,backgrounds}

\usepackage{wasysym}
\usepackage{balance}

\newcommand{\CUTLASS}{{CUTLASS C++}\xspace}
\newcommand{\Name}{{Tawa}\xspace}
\newcommand{\aref}{\texttt{aref}\xspace}
\newcommand{\aempty}{\texttt{empty}\xspace}
\newcommand{\afull}{\texttt{full}\xspace}
\newcommand{\aput}{\texttt{put}\xspace}
\newcommand{\aget}{\texttt{get}\xspace}
\newcommand{\aconsumed}{\texttt{consumed}\xspace}

\newcommand{\showcomments}{yes}
\newcommand\fixme[1]{
    \ifthenelse{\equal{\showcomments}{yes}}{\textcolor{red}{[#1]}}{\ignorespaces}
}
\newcommand\revise[1]{
    \ifthenelse{\equal{\showcomments}{yes}}{\textcolor{red}{#1}}{\ignorespaces}
}
\newcommand\hz[1]{
    \ifthenelse{\equal{\showcomments}{yes}}{\textcolor{red}{\small [hz: #1~]}}{\ignorespaces}
}
\newcommand\vg[1]{
    \ifthenelse{\equal{\showcomments}{yes}}{\textcolor{purple}{\small [vg: #1~]}}{\ignorespaces}
}
\newcommand\zz[1]{
    \ifthenelse{\equal{\showcomments}{yes}}{\textcolor{blue}{\small [zz: #1]}}{\ignorespaces}
}

\usepackage{xcolor}

\newcommand{\proposal}[3][]{%
  {\color{green!50!black}%
    \if\relax\detokenize{#1}\relax
    \else
      \leavevmode\raisebox{0.8ex}{\scriptsize\textnormal{[#1]}}\kern0.3em
    \fi
    #3%
  }%
}

\begin{document}
\title{\Name: Automatic Warp Specialization for\\Modern GPUs with Asynchronous References}

\author{
\IEEEauthorblockN{Hongzheng Chen\IEEEauthorrefmark{1}}
\IEEEauthorblockA{\textit{Cornell University}, USA\\
hzchen@cs.cornell.edu}
\\
\IEEEauthorblockN{Evghenii Gaburov}
\IEEEauthorblockA{\textit{NVIDIA}, USA\\
egaburov@nvidia.com}
\\
\IEEEauthorblockN{Jason Knight}
\IEEEauthorblockA{\textit{NVIDIA}, USA\\
jaknight@nvidia.com}
\and
\IEEEauthorblockN{Bin Fan}
\IEEEauthorblockA{\textit{NVIDIA}, USA\\
binf@nvidia.com}
\\
\IEEEauthorblockN{Masahiro Masuda}
\IEEEauthorblockA{\textit{NVIDIA}, USA\\
mmasuda@nvidia.com}
\\
\IEEEauthorblockN{Zhiru Zhang}
\IEEEauthorblockA{\textit{Cornell University}, USA \\
zhiruz@cornell.edu}
\and
\IEEEauthorblockN{Alexander Collins}
\IEEEauthorblockA{\textit{NVIDIA}, United Kingdom\\
acollins@nvidia.com}
\\
\IEEEauthorblockN{Matthew Brookhart}
\IEEEauthorblockA{\textit{NVIDIA}, USA\\
mbrookhart@nvidia.com}
\\
\IEEEauthorblockN{Vinod Grover}
\IEEEauthorblockA{\textit{NVIDIA}, USA\\
vgrover@nvidia.com}
\and
\IEEEauthorblockN{Bastian Hagedorn}
\IEEEauthorblockA{\textit{NVIDIA}, Germany\\
bhagedorn@nvidia.com}
\\
\IEEEauthorblockN{Chris Sullivan}
\IEEEauthorblockA{\textit{NVIDIA}, USA\\
chrsullivan@nvidia.com}
}


\maketitle
\begingroup\renewcommand\thefootnote{\IEEEauthorrefmark{1}}
\footnotetext{Work partially done while interning at NVIDIA.}
\endgroup

\input{sections/0-abstract}
\input{sections/1-introduction}
\input{sections/2-background}
\input{sections/3-compiler}
\input{sections/4-optimization}

\input{sections/5-experiments}

\input{sections/6-discussion}
\input{sections/7-conclusion}
\input{sections/ack}

\IEEEtriggeratref{42}
\bibliographystyle{IEEEtran}
\bibliography{cgo26}


\end{document}

%% file: sections/0-abstract.tex
\begin{abstract}

Modern GPUs feature specialized hardware units that enable high-performance, asynchronous dataflow execution. However, the conventional SIMT programming model is fundamentally misaligned with this task-parallel hardware, creating a significant programmability gap. While hardware-level warp specialization is the key to unlocking peak performance, it forces developers to manually orchestrate complex, low-level communication and software pipelines---a process that is labor-intensive, error-prone, and unsustainable.
To address this challenge, we present \Name, an automated compiler that systematically generates high-performance, warp-specialized code from a high-level, tile-based program.
Central to our approach is a novel IR abstraction, asynchronous references (\aref), which expresses warp-level communication without exposing low-level hardware details. Using this abstraction, \Name automatically partitions programs into producer-consumer roles and manages the intricate dataflow pipeline, relieving developers of invasive kernel rewriting.
Evaluation on NVIDIA H100 GPUs across representative LLM kernels shows that \Name delivers high hardware utilization, achieving up to 1.1$\times$ speedup over highly optimized cuBLAS GEMM kernels. For attention workloads, \Name attains 1.2$\times$ speedup over Triton and matches the performance of the hand-optimized \CUTLASS FlashAttention-3 kernel with far less programming effort.
\end{abstract}


%% file: sections/1-introduction.tex
\section{Introduction}
\label{sec:intro}
Modern GPU architectures have evolved from homogeneous processors into complex heterogeneous systems, integrating multiple specialized hardware units to maximize performance. Alongside traditional CUDA cores, modern GPUs integrate dedicated engines such as Tensor Cores for dense matrix computations and the Tensor Memory Accelerator (TMA) for asynchronous data movement. This architectural design enables a fine-grained, asynchronous dataflow execution model where data movement can be explicitly orchestrated and overlapped with computation. Exploiting this capability is critical for latency-sensitive workloads such as Large Language Models (LLMs), whose efficiency is often limited more by data orchestration than by raw FLOPs~\cite{amir2024memorywall,chen2024understanding,pope2023scalellmtpu,ivanov2021datamovement,chen2024slapo}.

However, the prevailing programming paradigm for GPUs, Single-Instruction Multiple-Thread (SIMT), which underpins CUDA, is fundamentally misaligned with this heterogeneous, task-parallel reality. The SIMT model assumes that all threads within a group (a warp) execute the same instructions on different data, excelling at problems with uniform control flow. This model is ill-suited for managing diverse hardware units that must be driven by different, concurrent tasks. For example, one task might involve generating addresses and initiating a bulk copy with the TMA, while another performs a matrix multiplication on the Tensor Cores. This creates a paradigm mismatch: the hardware is capable of asynchronous, task-parallel execution, but the software model is built for synchronous, data-parallel execution.

To resolve this conflict, modern architectures like NVIDIA's Hopper and Blackwell GPUs introduce hardware-level \emph{warp specialization}~\cite{hopper,blackwell}. This mechanism reflects the hardware’s task-parallel nature by allowing different groups of threads (warps) to take on distinct roles. For instance, producer warps can be dedicated to managing data transfers via the TMA, feeding data to consumer warps that execute computations on the Tensor Cores. While this new model unlocks the hardware's potential, programming for it remains difficult because the surrounding software ecosystem and abstractions are still rooted in the SIMT paradigm. This programmability gap makes developing high-performance kernels a labor-intensive endeavor, often requiring over a thousand lines of hand-optimized code even for a basic GEMM kernel. This difficulty manifests in three primary challenges:

\textbf{Challenge 1: Coordinating concurrent warp roles.}
Because the SIMT model lacks first-class abstractions for assigning heterogeneous roles, developers must manually restructure kernels, encapsulating role-specific logic within complex conditional branches. They must then orchestrate the execution of these specialized warps to ensure that producer and consumer tasks remain correctly synchronized and efficiently overlapped. This manual process is inherently fragile. Even modern, high-level domain-specific languages (DSLs) like Triton~\cite{tillet2019triton} retain SIMT-centric abstractions, which preclude the explicit assignment of warp roles and thus cannot naturally express warp-specialized programs.

\textbf{Challenge 2: Low-level communication management.}
CUDA does not currently expose C++-level interfaces for TMA and transaction barrier (mbarriers) operations. Consequently, programmers must rely on inline PTX to configure TMA descriptors, initialize and manipulate mbarriers, and orchestrate synchronization. This requires reasoning explicitly about arrival counts, barrier phases, and shared-memory aliasing, where subtle errors can easily lead to deadlock. The difficulty is exacerbated when multiple mbarriers and input parameters are involved. Such low-level orchestration undermines code readability and substantially complicates debugging and maintenance. High-level libraries like CUTLASS~\cite{cutlass} and ThunderKittens~\cite{spector2025thunderkittens} offer pre-written kernels but still require users to carefully compose them to satisfy the underlying communication requirements.

\textbf{Challenge 3: Resource allocation and pipeline orchestration.}
Effectively overlapping the work of specialized producer and consumer warps necessitates the manual implementation of deep software pipelines. Developers must explicitly manage multi-buffering schemes in shared memory, partition kernels into prologue, main body, and epilogue stages, and carefully balance the allocation of hardware resources like registers and shared memory to respect occupancy limits. These design choices are tightly coupled and create a vast, complex optimization space. For example, increasing the pipeline depth affects shared memory capacity, which in turn constrains thread block size and impacts TMA descriptor layouts. Navigating this space requires repeated, labor-intensive kernel refactoring, making performance tuning both difficult and unsustainable.

To address these challenges, we propose \Name\footnote{A shorthand for \underline{T}ask-\underline{A}ware \underline{W}arp Specialization with \underline{A}ref.
The implementation has been upstreamed to OpenAI Triton: \href{https://github.com/triton-lang/triton/tree/aref_auto_ws}{https://github.com/triton-lang/triton/tree/aref\_auto\_ws}, with the first PR available at \href{https://github.com/triton-lang/triton/pull/6288}{\#6288}.}
an automated compiler that systematically analyzes computation graphs generated from a tile-based DSL, partitions programs into warp-specialized components, and manages both communication and pipelining. Our contributions are as follows:
\begin{itemize}
\item We propose \emph{asynchronous references} (\aref), an abstraction for warp-level communication with formal guarantees. It captures the intent of asynchronous data movement without exposing hardware-specific intrinsics, enabling a clean and portable representation at the IR level.
\item We develop \Name, the first fully automated compilation flow for warp specialization on modern NVIDIA GPUs. \Name operates on unmodified, annotation-free Triton programs, performs task-aware partitioning across warp groups, applies multi-granularity pipelining adapted to diverse kernel structures, and emits high-performance PTX code for GPU execution.
\item We evaluate \Name on an H100 GPU using representative LLM kernels. Our approach delivers high hardware utilization with concise kernel code, matching or surpassing hand-optimized baselines. For GEMM, \Name achieves up to 1.1$\times$ speedup over cuBLAS, and for attention kernels, it attains 1.2$\times$ speedup over Triton and matches the performance of the hand-optimized \CUTLASS FlashAttention-3~\cite{shah2024fa3} kernel.
\end{itemize}

%% file: sections/2-background.tex
\section{Background and Related Work}
\label{sec:background}
In this section, we first discuss the modern GPU architecture and briefly summarize the existing GPU programming models.

\subsection{Modern NVIDIA GPU Architecture}
\begin{figure}[t]
\centering
\includegraphics[width=\linewidth]{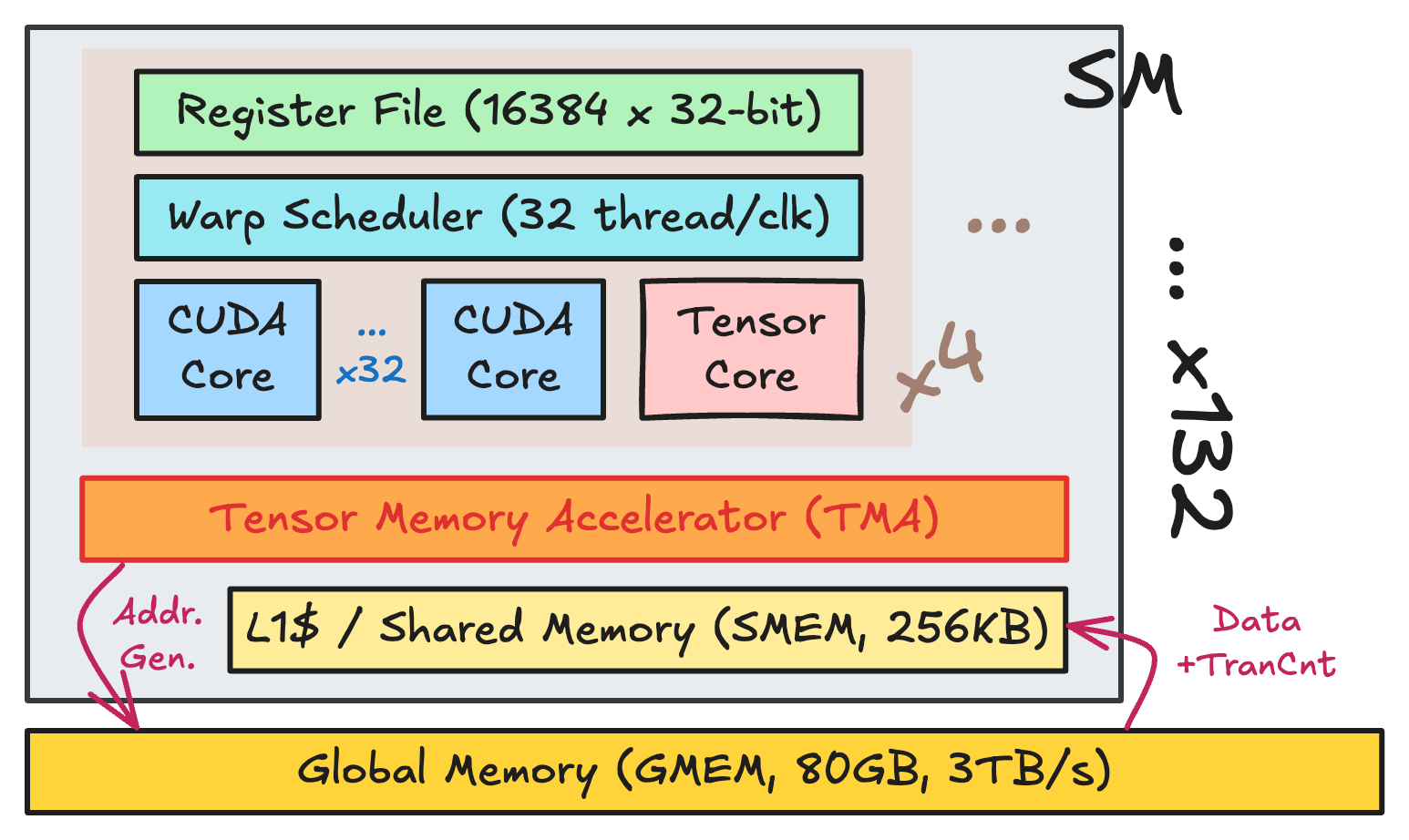}
\caption{A simplified NVIDIA H100 SXM5 GPU architecture.}
\label{fig:hopper}
\end{figure}
\input{code/example}
As shown in Fig.~\ref{fig:hopper}, the Streaming Multiprocessor (SM) forms the fundamental compute unit of modern GPUs, encapsulating both computation and memory resources. Each SM integrates a large register file, a warp scheduler that can issue instructions to one warp (32 threads) per clock cycle, and multiple CUDA cores that handle general-purpose arithmetic. Additionally, SMs include specialized Tensor Cores for accelerating matrix computations and a memory block that serves as software-managed shared memory (SMEM).
A thread block, or Cooperative Thread Array (CTA), is scheduled on an SM and shares its registers and SMEM.
Recent GPU generations, such as Hopper, have increased their reliance on Tensor Cores as the dominant source of compute throughput, where 90\% of the total compute throughput is contributed by Tensor Cores~\cite{hopper}. As such, achieving high performance on modern GPUs is to fully saturate these Tensor Cores with sufficient and timely data.

To this end, Hopper introduces two key hardware innovations aimed at improving Tensor Core utilization: the Tensor Memory Accelerator (TMA) and Warp Group Matrix Multiply-Accumulate (WGMMA). The TMA facilitates hardware-managed, multidimensional data transfers from global memory (GMEM) to shared memory, offloading the address generation and transfer coordination tasks from software to a dedicated unit.
To further enable this overlap, Hopper extends the programming model with asynchronous transaction barriers, which allow threads to synchronize not just based on arrival but also on the volume of data exchanged. Each transaction carries a byte count, and the barrier only releases threads once both all producers have arrived and the expected total transaction count is reached. This mechanism is especially powerful in coordinating asynchronous memory copies and collective data exchanges across thread blocks.

In parallel, WGMMA introduces warp-group-level matrix multiply operations, allowing multiple warps (4 warps as one warp group) to cooperatively execute large-tile GEMM kernels in an asynchronous and efficient manner. Together, these mechanisms provide a foundation for warp specialization, where different warps within an SM assume distinct roles in data communication and computation, thereby improving throughput for LLM workloads~\cite{shah2024fa3,ye2025flashinfer,spector2025thunderkittens}.
These trends illustrate a broader architectural shift toward asynchronous dataflow execution, a direction continued in subsequent architectures such as Blackwell~\cite{blackwell}.

\subsection{Existing GPU Programming Models}
CUDA is the dominant interface for GPU programming~\cite{nickolls2008cuda}, but as architectures evolve, programming complexity has increased substantially. NVIDIA's \CUTLASS~\cite{cutlass} library provides templated kernels for GEMM and related computations, but leveraging it to write neural network operators requires intricate configuration and hand-tuning. To reduce engineering effort for domain workloads, specialized libraries have emerged. ThunderKittens~\cite{spector2025thunderkittens} introduces a compact C++ interface based on 16$\times$16 tile abstractions, offering performance-competitive GEMM and attention kernels. FlashInfer~\cite{ye2025flashinfer} focuses on efficient attention and KV-cache management with JIT specialization to support large-scale LLM inference.
These systems simplify access to high-performance kernels but remain library-centric: they provide optimized operator templates rather than a general-purpose strategy for different applications, which may cause extensive maintenance issues for hand-tuning performance for different sizes of matrices.

To further raise the abstraction level, domain-specific compiler infrastructures have been proposed. TVM~\cite{chen2018tvm} pioneered an end-to-end compilation framework with automated scheduling~\cite{zheng2020ansor,shao2022metaschedule}, while Relax~\cite{lai2025relax} extends it with a graph-level IR for dynamic workloads. Fireiron~\cite{hagedorn2020fireiron} also leverages a schedule language for specifying the tiling and data movement, and Graphene~\cite{hagedorn2023graphene} further enhances this idea by providing an IR for mapping tensors with various layouts to Tensor Core instructions. While powerful, these systems were developed prior to Hopper and do not directly exploit TMA and hardware warp specialization, leaving significant performance untapped.
Similarly, lots of efforts on GPU warp specialization ~\cite{wapman2023harmoniccuda,bauer2011cudadma,bauer2014singe} were proposed before the Hopper architecture, which do not effectively use the latest TMA features.
WASP~\cite{crago2024wasp} considers GPUs with TMA but requires architectural modifications and runs solely on a simulator. In contrast, \Name targets commercial NVIDIA GPUs and demonstrates measurable speedups on real workloads.

Motivated by the need for finer control, a new class of tile-based kernel languages has emerged. Triton~\cite{tillet2019triton}, built on MLIR~\cite{chris2021mlir}, allows programmers to write efficient kernels in Python while abstracting tiling and memory management. Despite its success in democratizing kernel development, Triton, before this work, still lacks native abstractions for warp specialization and instead relies on Ampere-style software pipelining scheme with \texttt{cp.async}, leaving a performance gap on the Hopper architecture.
Recent work introduces explicit APIs for dataflow scheduling and pipeline composition.
Cypress\cite{yadav2025cypress} proposes a task-based programming model that abstracts complex data movement and synchronization through logical computation descriptions and mapping specifications. However, programmers are still required to manually manage the multi-level hierarchy of threads, warps, and SMs, along with the corresponding data decomposition.
Gluon~\cite{gluon} proposes an IR for Triton that adds layout abstractions but still requires programmers to explicitly incorporate communication operations.
These approaches provide expert users with expressive control, but they continue to impose significant responsibility for synchronization and resource management, which limits programmer productivity.

Additionally, TileLang~\cite{wang2025tilelang,cheng2025pipethreader} is a new tile-based GPU kernel language with built-in warp specialization support. It implicitly schedules operations to form pipelines, but without a formal correctness guarantee and with limited control over fine-grained MMA pipelines. In contrast, \Name formally defines asynchronous communication semantics via the \aref abstraction (\S\ref{sub:aref}), treating communication as a first-class IR construct. This design ensures that the compiler generates correct warp-specialized code by construction, preventing deadlocks such as premature data access in \aref objects. \Name’s multi-granularity pipelining further leverages \aref to guarantee correctness across diverse communication patterns. Importantly, \Name requires no user annotations or code modifications, whereas TileLang still requires marked pipelined loops (\texttt{T.pipelined}) and explicit copy operations (\texttt{T.copy}) to define pipeline stages.

Taken together, these trends highlight a growing need for raising the level of abstraction. Users should not cope with low-level hardware details or manually orchestrate warp-group execution.
\Name bridges these gaps with a fully automated compiler equipped with \aref.


%% file: code/example.tex
\begin{figure*}[t]
\begin{subfigure}[b]{0.17\linewidth}
\centering
\includegraphics[width=\linewidth]{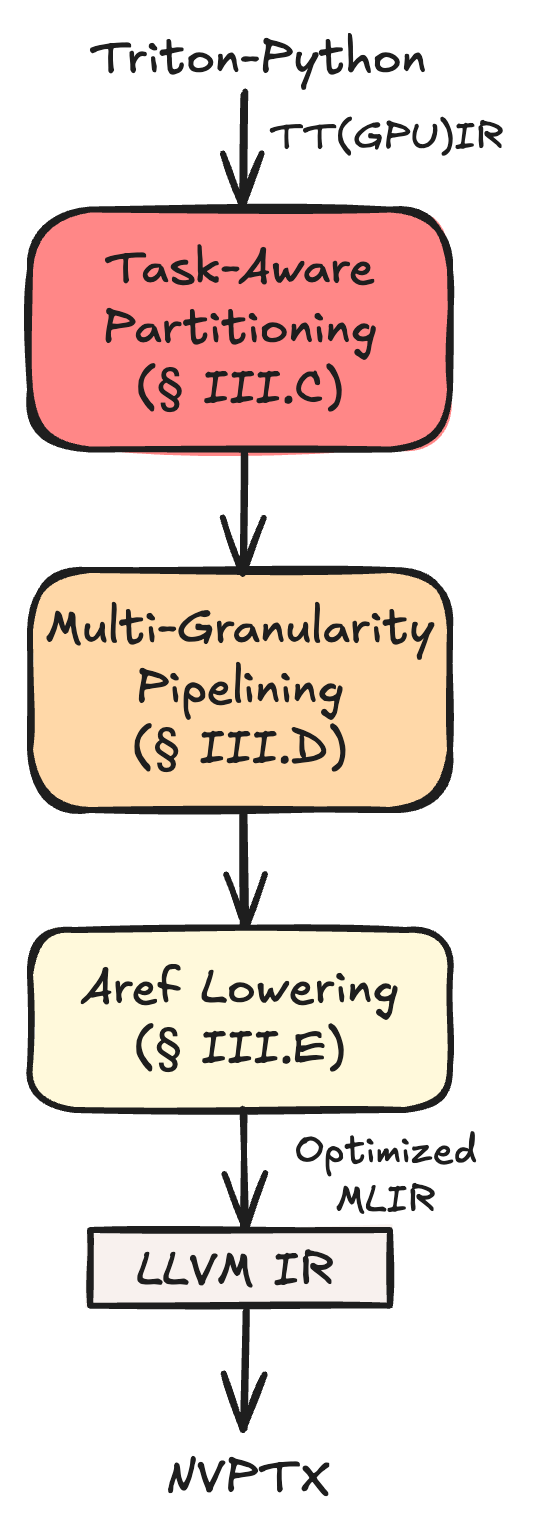}
\caption{\Name compilation}
\label{subfig:overview}
\end{subfigure}
\begin{subfigure}[b]{0.42\linewidth}
\begin{minted}[linenos,
               fontsize=\scriptsize,
               xleftmargin=1.8em,
               escapeinside=||,
               highlightlines={16,17,20},
               numbersep=5pt]{python}
@triton.jit
def matmul(a_desc, b_desc, c_ptr,
    M, N, K,
    stride_cm: tl.const, stride_cn: tl.const,
    Mt: tl.const, Nt: tl.const, Kt: tl.const):
 pid = tl.program_id(axis=0)
 num_pid_m = tl.cdiv(M, Mt)
 pid_m = pid % num_pid_m
 pid_n = pid // num_pid_m
 o_am = pid_m * Mt
 o_bn = pid_n * Nt
 o_k = 0
 acc = tl.zeros((Mt, Nt), dtype=tl.float32)
 for k in range(0, tl.cdiv(K, Kt)):
  # Data loading
  a = tl.tma_load(a_desc, [o_am, o_k], [Mt, Kt])
  b = tl.tma_load(b_desc, [o_bn, o_k], [Nt, Kt])
  # Computation
  acc = tl.dot(a, b.T, acc=acc)
  o_k += Kt
 offs_cm = pid_m * Mt + tl.arange(0, Mt)
 offs_cn = pid_n * Nt + tl.arange(0, Nt)
 c_ptrs = c_ptr + stride_cm * offs_cm[:, None] \
                + stride_cn * offs_cn[None, :]
 tl.store(c_ptrs, acc)
\end{minted}
\centering
\includegraphics[width=0.9\linewidth]{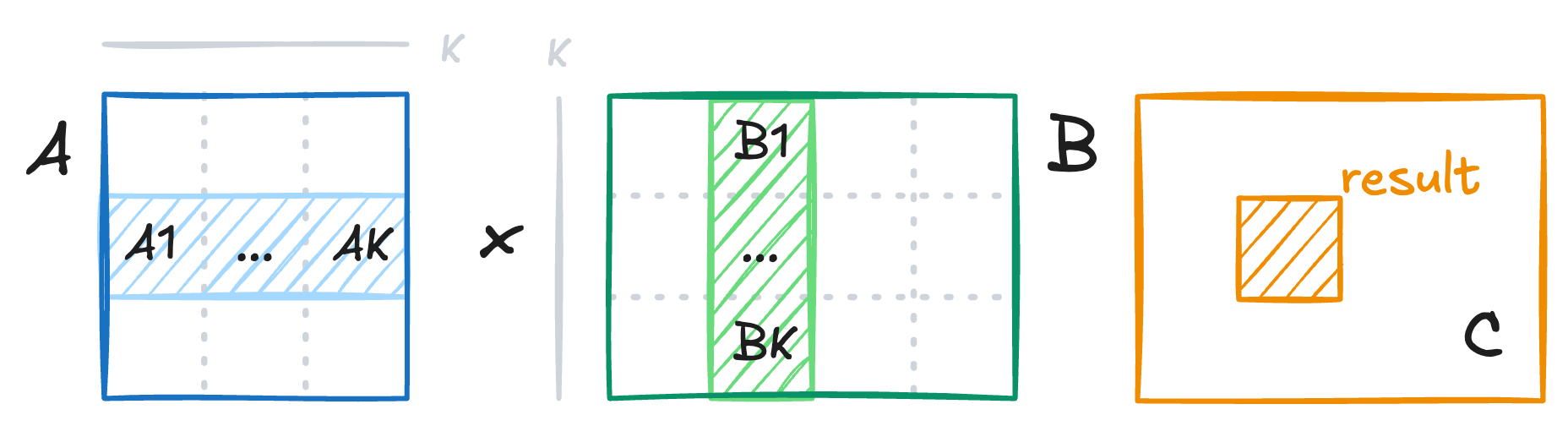}
\caption{A simplified Triton GEMM program}
\label{subfig:triton}
\end{subfigure}
\begin{subfigure}[b]{0.4\linewidth}
\begin{minted}[linenos,
               fontsize=\scriptsize,
               xleftmargin=1.8em,
               escapeinside=||,
               highlightlines={6-8,16,24,27},
               numbersep=5pt]{java}
module attributes {"num-warps" = 8 : i32} {
 tt.func public @matmul(
    %arg0: !tt.ptr<i8>, %arg1: !tt.ptr<i8>,
    %arg2: !tt.ptr<f16>, %arg3: i32,
    %arg4: i32, %arg5: i32) {
  %ABaref = tawa.create_aref : ()
   -> tensor<2x!tawa.aref<tuple
    <tensor<128x64xf16>, tensor<128x64xf16>>>
  // ... more initialization
  tawa.warp_group {
   scf.for %k = %c0 to %K step %c1
     iter_args(%arg8 = %c0) {
    %ta = tt.tma_load %arg0[%offset_am, %arg8]
    %tb = tt.tma_load %arg1[%offset_bn, %arg8]
    %extracted = %ABaref[%k mod %cst_D]
    tawa.put(%extracted, %ta, %tb)
    %kI = arith.addi %arg8, %c64
    scf.yield %kI : i32
  }} {partition = 0 : i32}
  tawa.warp_group {
   %tc = scf.for %k = %c0 to %K step %c1
     iter_args(%acc = %cst_0) {
    %extracted = %ABaref[%k mod %cst_D]
    %ta, %tb = tawa.get(%extracted)
    %part_ret = tt.dot_wait %ta, %tb, %acc
    %res = tt.dot_wait %part_ret {pendings = 1}
    tawa.consumed(%extracted)
    scf.yield %res : tensor<128x128xf32>
   }
   // epilogue: calculate C address and store
   tt.store %c_ptr, %tc
  } {partition = 1 : i32}
}}
\end{minted}
\caption{Highly simplified \Name MLIR code ($D=2$)}
\label{subfig:mlir}
\end{subfigure}
\caption{Example \Name compilation flow from the Triton frontend to the internal MLIR representation. \texttt{tl} is the shorthand for the Triton language. \texttt{tt} is the Triton MLIR dialect. Code snippets are simplified for demonstration purposes.}
\label{fig:overview}
\end{figure*}

%% file: sections/3-compiler.tex
\section{\Name compiler}
In this section, we first give an overview of the \Name compiler and discuss details of the asynchronous reference abstraction and the compilation process.

\subsection{Overview}
Fig.~\ref{subfig:overview} illustrates the overall design of \Name. Since Triton~\cite{tillet2019triton} has become the de facto backend for PyTorch~\cite{paszke2019pytorch,pytorch20} and is widely adopted for writing efficient GPU kernels, we select Triton as the frontend interface of our compiler. Programmers write kernels in Triton-Python with tiled computation and TMA communication in a CTA (Fig.~\ref{subfig:triton}), which are first translated into the standard Triton-MLIR representation. Building upon this infrastructure, \Name introduces a series of compiler passes that automate the generation of warp-specialized programs directly from high-level Python code.
Most importantly, we require \emph{no} modifications to existing Triton code; users can simply set \texttt{enable\_warp\_specialization=True} when calling the Triton kernel, and it only takes effect on Hopper and newer GPU architectures.

To address the challenges outlined in \S~\ref{sec:intro}, \Name applies three key transformations:
\textbf{(1) Task-aware partitioning (\S~\ref{sub:partition})}. We introduce a novel algorithm that partitions the program into producer and consumer warp groups, ensuring correct role assignment and communication boundaries, directly addressing \textbf{Challenge 1}.
\textbf{(2) Asynchronous reference abstraction (\S~\ref{sub:aref})}. We design \aref, an intermediate representation that captures inter-warp dataflow explicitly. By expressing communication intent at the IR level, \aref allows the compiler to automatically insert and coordinate the necessary synchronization and data-movement operations, resolving \textbf{Challenge 2}.
We also propose a new \Name MLIR dialect to encode those \aref operations.
\textbf{(3) Multi-granularity software pipelining (\S~\ref{sub:sw_pipeline})}. On top of the \aref program, we apply a software pipelining pass that orchestrates overlap between communication and computation across multiple granularities, resolving \textbf{Challenge 3}.

The resulting program is shown in Fig.~\ref{subfig:mlir}. Once these passes are applied, the \aref representation is lowered to the standard Triton-MLIR dialect, after which the existing Triton compilation pipeline proceeds unchanged: the code is progressively lowered to LLVM IR and finally to PTX, ready for execution on NVIDIA GPUs.

\subsection{Asynchronous Reference}
\label{sub:aref}
\begin{figure}[!htbp]
\centering
\includegraphics[width=\linewidth]{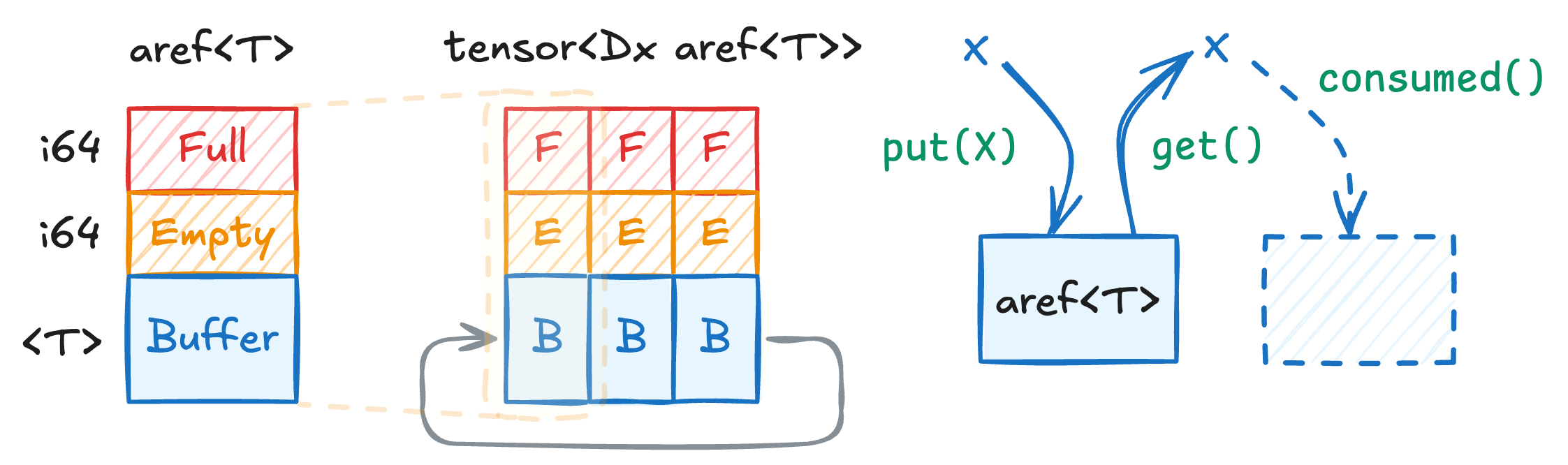}
\caption{\aref abstraction and associated operations.}
\label{fig:aref}
\end{figure}
To better model communication between warps, we introduce \emph{asynchronous reference}~\cite{bastian2025asyncgraphene},
or \aref, an IR abstraction that models a one-slot channel between producer and consumer on the GPU as shown in Fig.~\ref{fig:aref}.
It packages a data buffer together with two synchronization primitives implemented by hardware mbarriers, conventionally named \aempty and \afull. At any moment exactly one of these barriers encodes the state of the slot: when the \aempty mbarrier holds a credit, the slot may be written by the producer; when the \afull mbarrier holds a credit, the slot contains a published value ready to be read by the consumer.

\input{code/aref_semantics}
\input{code/aref_compilation}

The \aref interface exposes three operations: \aput, \aget, and \aconsumed, whose formal semantics are defined in Fig.~\ref{fig:aref_semantics}.
These semantics provide a rigorous foundation for lowering the operations to GPU-specific instructions, as discussed in \S~\ref{sub:lowering}.
\aput is the producer's publication step: it requires the slot to be empty (the \aempty mbarrier $E$ holds a credit), writes the payload to the buffer, and flips the state to full by arriving on the full mbarrier $F$ with release ordering.
This transition makes the data visible to the consumer and prevents subsequent writes from reusing the slot prematurely.
\aget is the consumer's acquisition step: it requires the slot to be full (the \afull mbarrier $F$ is observed), performs a read of the buffer into a program variable, and transitions the \aref into a borrowed state in which neither $E$ nor $F$ holds a credit, reflecting that the value is in use but the slot is not yet reusable.
\aconsumed closes the handshake: once the consumer no longer needs the value, it arrives on the empty mbarrier $E$, restoring the empty credit and enabling the next \aput; this induces the happens-before chain from producer's writes to consumer's reads and on to the producer's subsequent reuse.
The \aref therefore serves as a first-class IR value representing a concrete communication path that enforces correct ordering of data movement and computation across warp groups.

Compared to constructs such as the OpenCL \texttt{pipe}~\cite{opencl} or similar mechanisms in other accelerator programming models~\cite{oneapi,vitis,fang2025dato,chen2024allo,xiang2022heteroflow}, the \aref abstraction is specifically designed for GPU inter-warp communication. Its semantics are rooted in hardware-supported mbarrier synchronization, providing a clear ordering model. In addition to its synchronization semantics, \aref is also type-generic, supporting buffers of type \texttt{<T>}, which allows it to carry structured payloads such as tensors or tuples while reusing the same mbarriers. To facilitate deep pipelining, multiple \aref instances can be grouped into a cyclic buffer of depth $D$, effectively forming a ring structure that supports the storage and transfer of intermediate data across pipeline stages.

\subsection{Task-Aware Partitioning}
\label{sub:partition}
We begin by traversing the MLIR computation graph as in Fig.~\ref{subfig:comp_graph}, where each node is an MLIR operation and each edge captures the use-def dependencies. We take two steps of the partitioning process: partition annotation and loop distribution.

\subsubsection{Partition Annotation}
We perform a backward traversal along the use-def chains starting at the kernel's side-effecting sinks (e.g., \texttt{Store}). During the traversal, we attach a semantic tag to every node by inspecting its effects. Nodes that contribute to address computation, such as the pointer arithmetic for data transfers, are marked as \emph{iteration statements} (\textcolor{orange}{orange} edges). Nodes that transform or consume a tile for actual computation, such as WGMMA, are marked as \emph{tile statements} (\textcolor{blue}{blue} edges).
These tags make the high-level intent of each region explicit. Because iteration statements are often scattered throughout the IR, semantic tagging is necessary to recover producer-related operations.
For example, in Fig.~\ref{subfig:triton}, the address \texttt{o\_k} calculation in L20 is separated with the \texttt{TMALoad} operation in L16-17.

After tagging, \Name constructs the partitions. A partition is a dependency-closed subgraph that represents one warp role and will be executed by one warp group. To determine these subgraphs, \Name identifies the cut that separates producer behavior from consumer behavior. All iteration statements, together with the \texttt{TMALoad} operations they dominate, are grouped into the producer partition. All tile statements and their dependents form the consumer partition. The graph is then closed under dependencies, so each partition contains all operations required for correctness. If a node is used by both partitions, for example an address calculation that contributes to both a load and a store, \Name duplicates that computation. This guarantees that each partition is self-contained, and it avoids creating cross-partition data dependencies that would violate warp specialization constraints.

On Hopper, the typical configuration is two partitions. The producer partition corresponds to the load warp group (WG0), and the consumer partition corresponds to the compute warp group (WG1), as illustrated in Fig.~\ref{subfig:aref_comp_graph}. The epilogue is attached to WG1 so that output writes occur exactly once.
For Blackwell and later architectures that provide more warp roles, \Name can create additional partitions. The same semantic analysis drives these refinements, and the graph cut is adjusted to form separate load, compute, reduction, or epilogue subgraphs. In all cases, partitions are built from semantic roles, and duplication is applied when required so that each warp group receives a complete and independent subgraph for execution.

\subsubsection{Loop Distribution}
After the operations are annotated, we need to construct the actual IR that divides the workload for each WG.
For each cross-partition edge (e.g., \texttt{TMALoad}$\to$\texttt{LocalAlloc} in Fig.~\ref{subfig:comp_graph}), we create an \aref tensor with size $D$, denoting a $D$-slot cyclic buffer.
$D$ is chosen to maximize overlap between TMA transfers and Tensor Core computation. A larger $D$ allows the producer to run further ahead and hide more TMA latency, while a smaller $D$ reduces buffer footprint and synchronization overhead.
We study the effect of $D$ in \S~\ref{sub:hyperparam}.
A \aput operation is created on the producer side that places each \texttt{TMALoad} result into an \aref slot.
The slot index is computed as the iteration $k$ modulo the buffer size $D$, allowing the buffer slots to be reused (L15, L23 in Fig.~\ref{subfig:mlir}).
Similarly, a \aget operation is created on the consumer side that gets the data from the same \aref slot.
After the data are used for computation in WGMMA, an additional \aconsumed operation is inserted at the end to indicate the slot has been freed.
As mentioned in \S~\ref{sub:aref}, \aref supports grouping multiple values in the buffer to share the mbarrier.
Therefore, a straightforward optimization when creating \aref is to check whether the loaded tensors are used in the same WGMMA.
If so, we can put a tuple of tensors in the \aref buffer and share the channel for communication between WGs.
The resulting computation graph is shown in Fig.~\ref{subfig:aref_comp_graph}.
Since \aref has already carried buffers, the \texttt{LocalAlloc} operations can also be eliminated but to directly read from the \aref buffer.

To make both partitions independently progress through the original loop nest, we also need to perform loop distribution around the cut.
Two WG regions are created in the IR as shown in Fig.~\ref{subfig:mlir} (L10-32).
The \texttt{scf.for} loop is cloned so that WG0 and WG1 each carry an isomorphic loop over $k$. Iteration statements remain in WG0, and tile statements remain in WG1.
The inter-warp communication is done by \aref highlighted in the code (L16, L24).
Finally, the computed result is yielded from the for loop and stored back to global memory in the epilogue.
The resulting execution is the warp-specialized timeline in Fig.~\ref{subfig:ws_gemm}, which shows how TMA loads are overlapped with Tensor Core computation.

\subsection{Multi-Granularity Pipelining}
\label{sub:sw_pipeline}
After we have the warp-specialized program, we conduct further optimization on the compute warp group, where we introduce two pipelining mechanisms to make sure the hardware components are working in parallel.
Specifically, we consider a fine-grained pipeline that overlaps MMA address calculation and computation, and a coarse-grained pipeline that overlaps CUDA and Tensor Core computation.

\subsubsection{Fine-Grained Pipeline}
\begin{figure}[!htbp]
\centering
\includegraphics[width=\linewidth]{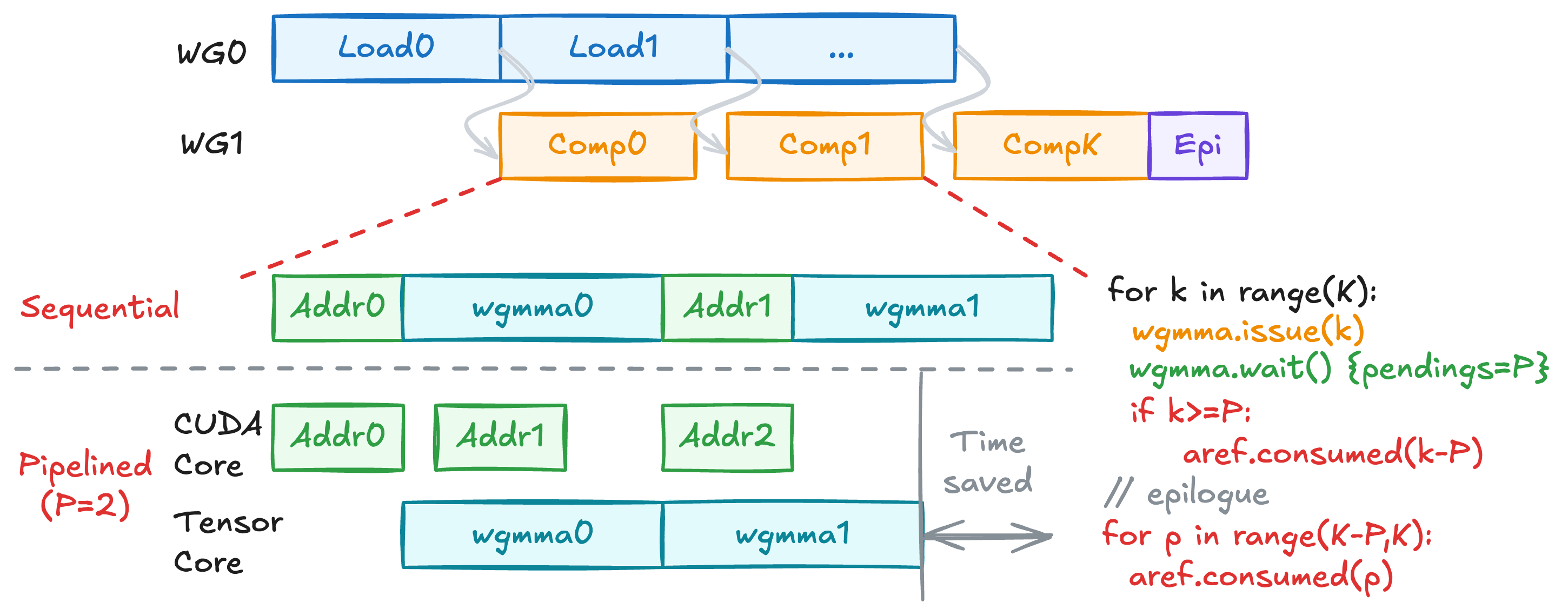}
\caption{Fine-grained pipeline overlapping MMA address calculation and computation. Actual latency not shown to scale.}
\end{figure}

Modern GPUs expose both CUDA Cores, which handle scalar and address computations, and Tensor Cores, which accelerate MMA operations; however, a na\"ive sequential execution forces these units to operate in sequence, such that address generation stalls tensor execution and vice versa, leading to poor utilization.
To alleviate this inefficiency, we introduce an automatic pipelining algorithm that overlaps address calculation on CUDA Cores with MMA execution on Tensor Cores through a bounded pipeline of depth $P$.
At each iteration, the compiler issues the next MMA instruction asynchronously (\texttt{WGMMA.issue(k)}), and only stalls when the number of outstanding operations reaches the maximum pipeline depth (\texttt{WGMMA.wait() \{pendings=P\}}).
As the $k$-th \texttt{wait} operation does not need to be done immediately after each MMA issue, enabling more MMA instructions to be in flight before waiting.
Once the pipeline is filled, the result reference from iteration $k-P$ is released (\texttt{aref.consumed(k-P)}), which maintains correctness by enforcing dataflow dependencies and freeing resources for reuse. After the final iteration, the compiler drains the pipeline in an epilogue by consuming the remaining pending operations. This design effectively decouples address computation and matrix multiplication, sustaining a continuous flow of work across both functional units, and effectively hides latency and reduces idle time compared to the sequential baseline.

\input{code/pipeline}

\subsubsection{Coarse-Grained Pipeline}
The fine-grained pipeline is most effective when the loop primarily consists of matrix multiplication operations. However, when additional computations are performed on CUDA Cores, these operations can also be overlapped with MMA.
Therefore, we further propose a coarse-grained three-stage producer-transform-(optional) consumer assembly line in Algorithm~\ref{alg:pipeline}.
Each loop iteration $j$ has a Tensor Core stage $T_j$ that produces an intermediate tile (e.g., a matmul fragment), a CUDA Core stage $C_j$ that transforms that intermediate (e.g., normalization, activation, reduction), and an optional downstream Tensor Core stage $U_j$ that consumes the transformed result (e.g., a second matmul/epilogue).
The prologue boots the pipeline by running $T_0$ to completion and immediately computing $C_0$.
Steady-state then overlaps $T_j$ with $C_{j-1}$ (and, if present, $U_{j-1}$): in each iteration we first enqueue and commit the Tensor Core work for the current tile, optionally enqueue and commit the downstream $U$ for the previous tile, then perform a precise completion wait on the dot-product group(s) that produced $T_{j-1}$'s results and run the CUDA Core transform $C_{j-1}$. The epilogue drains the last CUDA Core transform and, if $U$ is used, issues and retires $U_{N-1}$.
Correctness depends on two synchronization mechanisms: 
\textsc{DotWait}$(T_j-1)$ (and \textsc{DotWait}$(U_j-1)$) ensures compute-group completion at Tensor Core boundaries, while \textsc{MaybeArefGet}$(\cdot)$ are wrappers that perform the \aget operation only when a stage actually consumes a cross-WG \aref object.
When no external data needs to be read, the wrappers compile to no-ops, keeping the template minimal and avoiding unnecessary stalls.

From an MLIR computation graph, the schedule is constructed by several principled analyses followed by a mechanical synthesis step. Stage identification partitions the per-iteration subgraph into $T_j$, $C_j$, and (optionally) $U_j$ using dialect- and type-level cues: Tensor Core micro-tiles and their glue form $T$ (and possibly $U$ if a second tensor-core phase exists), whereas arithmetic, normalization, activation, reduction, or layout transformation that reads $T$'s outputs forms $C$. This yields a tiny intra-iteration DAG with edges $T_j\to C_j$ and, if present, $C_j\to U_j$.
An example of attention~\cite{vaswani2017transformer} is that the $QK^\mathrm{T}$ forms $T$, the softmax operation forms $C$, and the second GEMM $PV$ constructs $U$.
Finally, an \aref-use inspection marks any stage that reads data produced by a different warp-group; only for those marked stages will the wrappers \textsc{MaybeArefGet}$(\cdot)$ and 
\textsc{MaybeArefConsumed}$(\cdot)$ be emitted.

\subsection{\aref Lowering}
\label{sub:lowering}
The lowering of \aref transforms high-level asynchronous references into explicit synchronization and memory-transfer instructions that execute directly on the GPU. At the IR level, \texttt{create\_aref} declares a buffer and allocates the necessary mbarriers, serving as the root of a communication subgraph. The compiler then rewrites \aput, \aget, and \aconsumed operations into concrete instructions: \aput expands into an asynchronous TMA load guarded by \aempty/\afull barriers, \aget becomes a blocking wait on the corresponding \afull barrier, and \aconsumed signals buffer availability by arriving at the \aempty barrier. This pattern-matching ensures that once the entire subgraph is rewritten, the abstract \aref nodes can be safely erased, leaving only executable low-level synchronization and data-movement instructions.

A critical aspect of this transformation is correctness under pipelining, where multiple tiles are loaded, computed, and consumed in flight. To avoid deadlock from circular waits, the lowering introduces a parity mechanism: each operation alternates between two sets of barriers indexed by iteration parity. When parity toggles, a consumer may skip waiting if data has already been produced, and producers can reuse buffer slots without overwriting values still in use. This mechanism not only guarantees liveness but also enables efficient multi-buffering, allowing communication and computation to overlap across multiple stages. In this way, \aref lowering provides a systematic bridge between an abstract, declarative view of asynchronous dataflow and the precise TMA and mbarrier instructions required for high-performance GPU execution.

%% file: code/aref_semantics.tex
\begin{figure}[t]
\begin{mathpar}
\scalebox{0.9}{
\infer[Put]
{ \sigma(a).E = 1 }
{ \langle \sigma, \textsf{put}(a, v) \rangle
  \;\to\;
  \langle \sigma[\, a \mapsto \langle \textit{buf}=v,\, F=1,\, E=0 \rangle \,],\; \epsilon \rangle }
}

\scalebox{0.9}{
\infer[Get]
{ \sigma(a).F = 1 }
{ \langle \sigma, \textsf{get}(a) \rangle
  \;\to\;
  \langle \sigma[\, a \mapsto \langle \textit{buf}=\sigma(a).\textit{buf},\, F=0,\, E=0 \rangle \,],\; \sigma(a).\textit{buf} \rangle }
}

\scalebox{0.9}{
\infer[Consumed]
{ }
{ \langle \sigma, \textsf{consumed}(a) \rangle
  \;\to\;
  \langle \sigma[\, a \mapsto \langle \textit{buf}=\sigma(a).\textit{buf},\, F=0,\, E=1 \rangle \,],\; \epsilon \rangle }
}
\end{mathpar}
\caption{Operational semantics of the \aref operations. $\sigma$ is the store map from an \aref identifier $a$ to the actual structure $\sigma(a) = \langle \textit{buf}, F, E\rangle$, where $F$ is the \afull mbarrier flag and $E$ is the \aempty mbarrier flag. Initially $E=1$, $F=0$.}
\label{fig:aref_semantics}
\end{figure}

%% file: code/aref_compilation.tex
\begin{figure*}[t]
\begin{subfigure}[b]{0.29\linewidth}
\centering
\includegraphics[width=\linewidth]{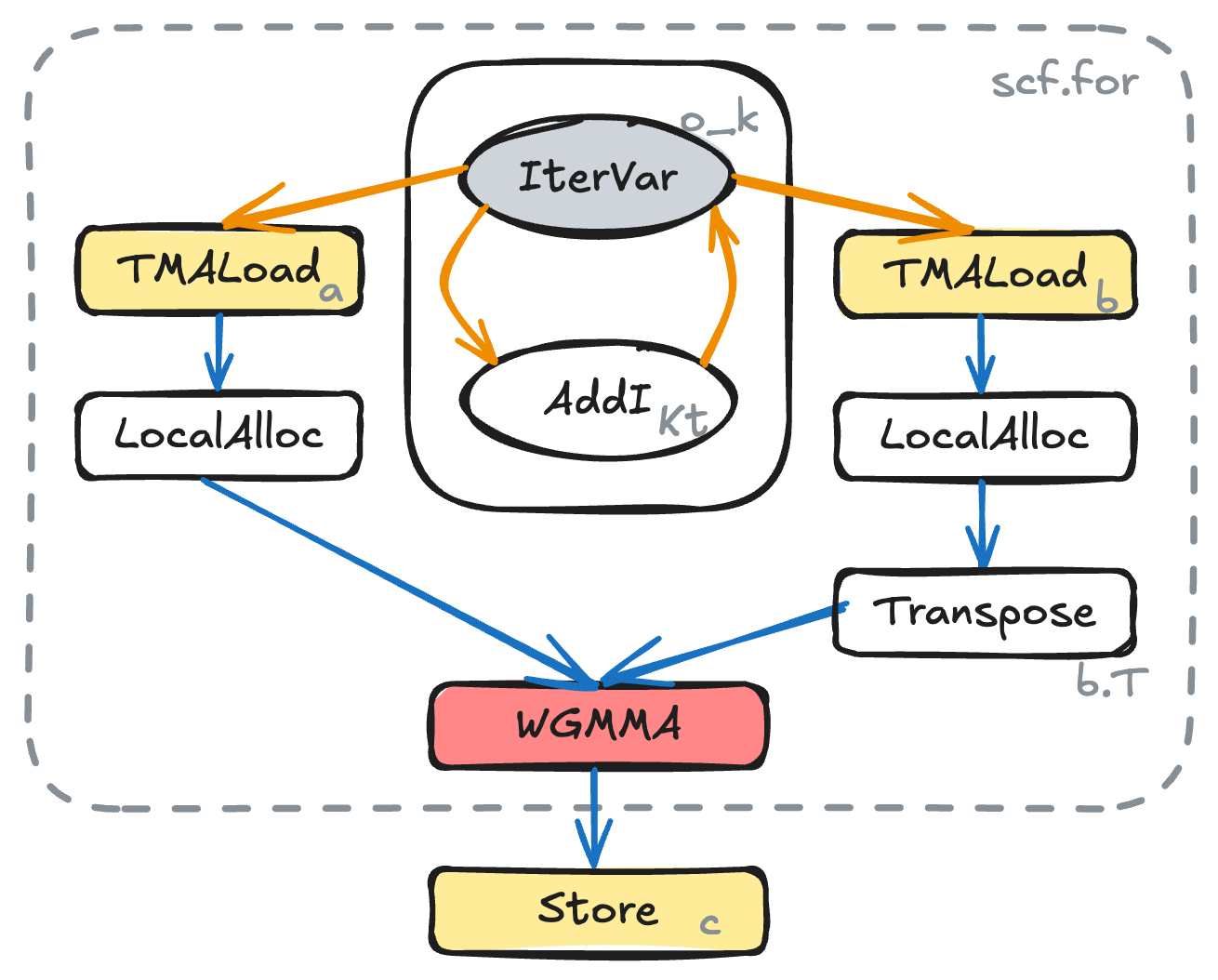}
\caption{Computation graph of Fig.~\ref{subfig:triton}.}
\label{subfig:comp_graph}
\end{subfigure}
\begin{subfigure}[b]{0.28\linewidth}
\centering
\includegraphics[width=\linewidth]{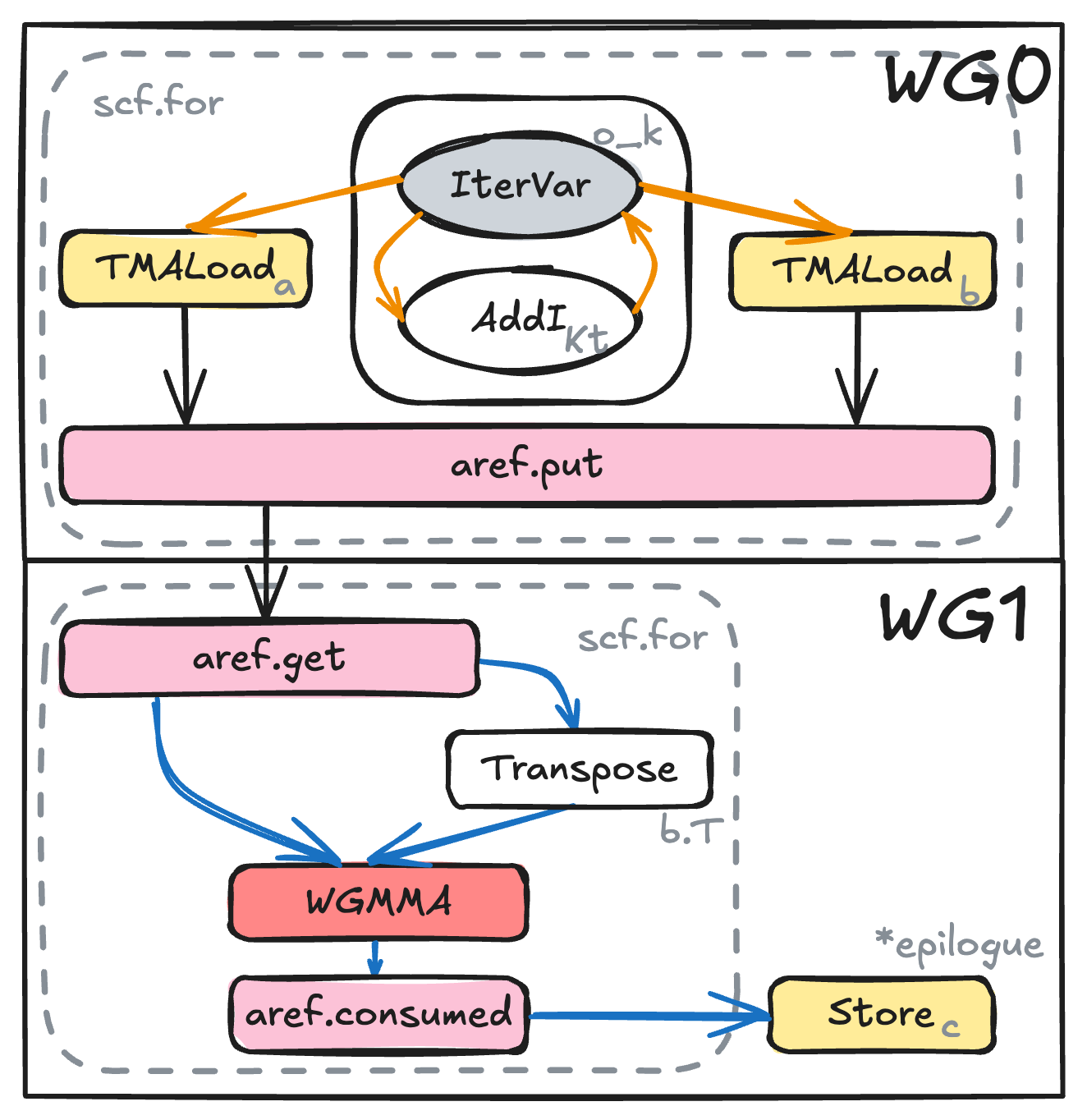}
\caption{Partitioned computation graph.}
\label{subfig:aref_comp_graph}
\end{subfigure}
\begin{subfigure}[b]{0.42\linewidth}
\centering
\includegraphics[width=\linewidth]{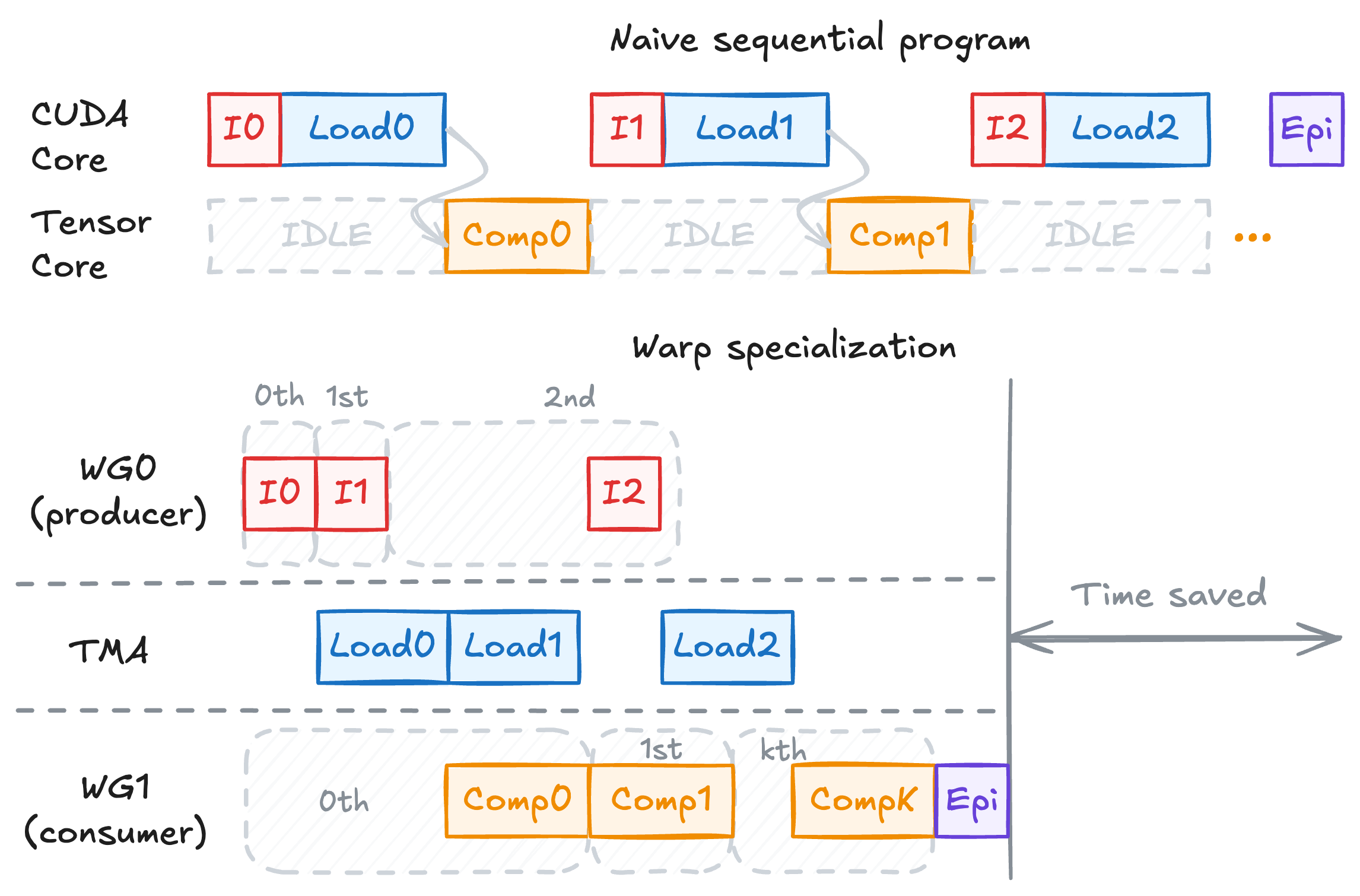}
\caption{Achieved warp specialization. Suppose $D=2$.}
\label{subfig:ws_gemm}
\end{subfigure}
\caption{Task-aware partition for the GEMM kernel in Fig.~\ref{subfig:triton}. Notice warp specialization is just one way of achieving the overlapping in (c), and the actual latency is not shown to scale.}
\label{fig:dag}
\end{figure*}

%% file: code/pipeline.tex
\begin{algorithm}[t]
\caption{Coarse-Grained CUDA and Tensor Core Pipelining}
\label{alg:pipeline}
\small
\KwIn{Tile count $N$; Boolean \texttt{USE\_U}}
\BlankLine
\textbf{Prologue}\;
\textsc{MaybeArefGet}$(T_{0})$\;
\textsc{IssueAndCommit}$(T_{0})$\;
\textsc{DotWait}$(T_{0})$ \tcp*[r]{materialize $T_{0}$ output}
\textsc{MaybeArefConsumed}$(T_{0})$\;
\textsc{Compute}$(C_{0})$\;

\BlankLine
\textbf{Steady state}\;
\For{$j \leftarrow 1$ \KwTo $N-1$}{
  \textsc{MaybeArefGet}$(T_{j})$;
  \textsc{IssueAndCommit}$(T_{j})$\;

  \If{\texttt{USE\_U}}{
    \textsc{MaybeArefGet}$(U_{j-1})$\;
    \textsc{IssueAndCommit}$(U_{j-1})$\;
  }

  \textsc{DotWait}$(T_{j-1})$ \tcp*[r]{ensure $T_{j-1}$ results are visible}
  \textsc{MaybeArefConsumed}$(T_{j-1})$;
  \textsc{Compute}$(C_{j-1})$\;

  \If{\texttt{USE\_U}}{
    \textsc{DotWait}$(U_{j-1})$\;
    \textsc{MaybeArefConsumed}$(U_{j-1})$
  }
}

\BlankLine
\textbf{Epilogue}\;
\textsc{DotWait}$(T_{N-1})$\;
\textsc{MaybeArefConsumed}$(T_{N-1})$\;
\textsc{Compute}$(C_{N-1})$\;
\If{\texttt{USE\_U}}{
  \textsc{MaybeArefGet}$(U_{N-1})$\;
  \textsc{IssueAndCommit}$(U_{N-1})$\;
  \textsc{DotWait}$(U_{N-1})$\;
  \textsc{MaybeArefConsumed}$(U_{N-1})$
}
\textsc{FinalEpilogueAndStore}()\;
\end{algorithm}

%% file: sections/4-optimization.tex
\section{Further Optimizations}
\label{sec:opt}
To further increase the Tensor Core utilization in warp-specialized scenarios, \Name further integrates additional optimizations that are incorporated into the compilation flow, ensuring that programs automatically benefit without requiring manual restructuring.
\begin{figure}[t]
\centering
\includegraphics[width=\linewidth]{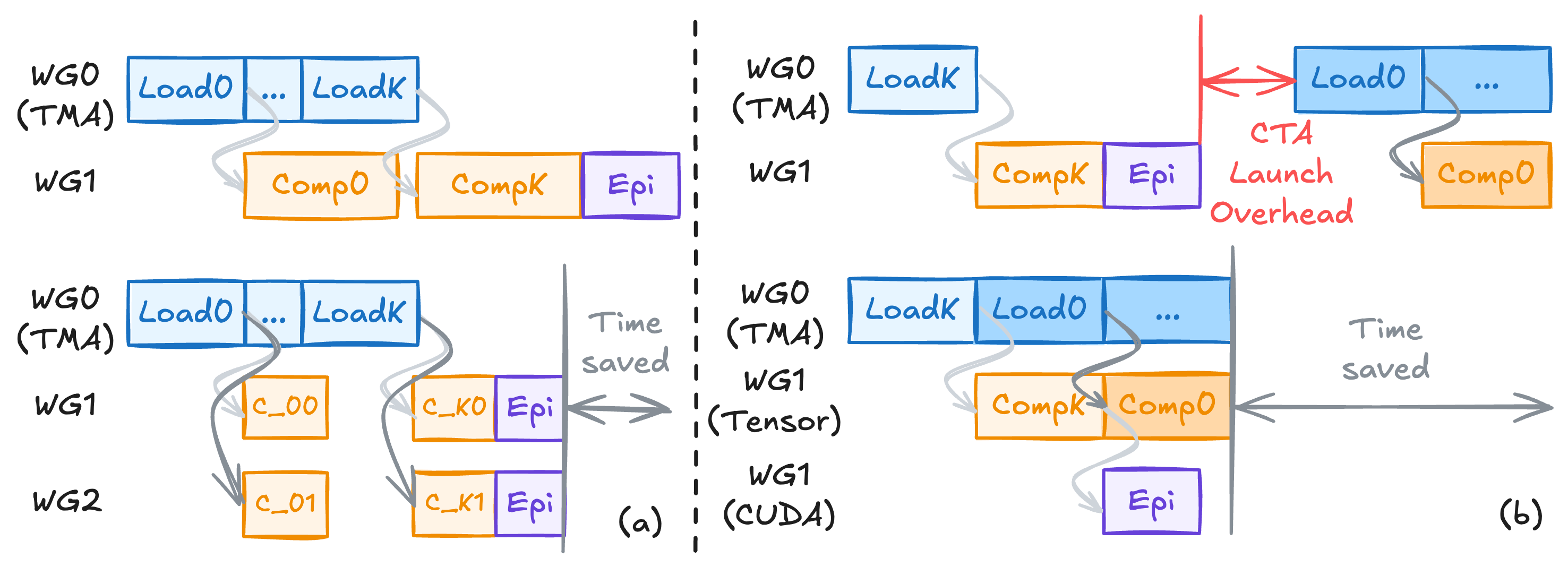}
\caption{(a) Cooperative warp groups; (b) Persistent kernel.}
\label{fig:further_opt}
\end{figure}

\subsection{Cooperative Compute Warp Groups}
A key limitation of warp-specialized execution is the restricted register budget available to each warp group, which constrains the maximum tile size and can limit overall compute intensity. The cooperative warp group optimization alleviates this bottleneck by enabling multiple warp groups to collaboratively compute the same tile. As illustrated in Fig.~\ref{fig:further_opt}a, two consumer warp groups (WG1 and WG2) jointly consume data produced by a TMA warp group (WG0). By pooling their registers, the cooperating warps can form larger tiles that increase arithmetic intensity, improve data reuse, and reduce memory traffic. Conceptually, this optimization reintroduces a form of standard warp parallelism that multiple warps performing the same computation on different fragments of the same tile, into the warp-specialized setting. \Name handles this transparently by informing the backend code generator of the cooperative mapping so that thread indices are consistently assigned. Importantly, no changes are needed to the \aref abstraction, since communication semantics remain unchanged.

\subsection{Persistent Kernels}
Another major source of inefficiency in GPU pipelines stems from repeated kernel launches, each introduces significant CTA launch overhead. Persistent kernels reduce this overhead by launching only as many CTAs as there are SMs and keeping them resident throughout execution. As depicted in Fig.~\ref{fig:further_opt}b, \Name transforms the execution model into a software-managed pipeline, where resident CTAs are responsible for iterating over tiles. This enables overlapping of TMA-driven loads, Tensor Core computation, and CUDA Core epilogues across iterations, eliminating relaunch latency and reducing idle time. \Name automatically inserts synchronization before each epilogue to preserve correctness. This optimization is especially effective in deep pipelines with large $K$ dimensions, where cumulative launch costs would otherwise dominate.


%% file: sections/5-experiments.tex
\section{Evaluation}
\label{sec:exp}
We first describe our experimental setting and then show the results on different benchmarks in this section.

\begin{figure*}[t]
    \centering
    \includegraphics[width=\linewidth]{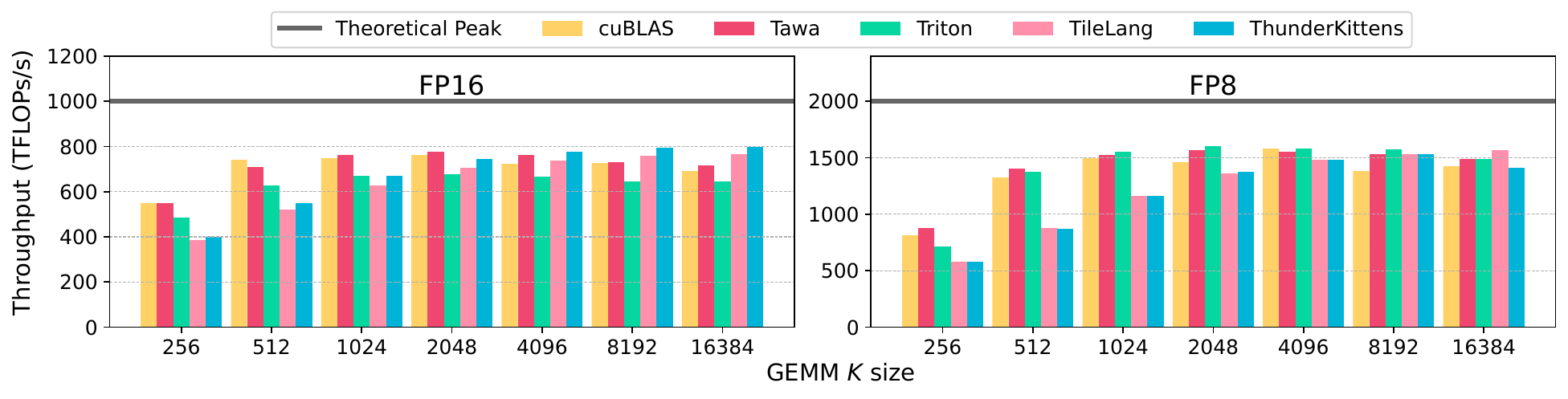}
    \caption{GEMM performance results for different frameworks. $M=N=8192$, and varied $K$.}
    \label{fig:gemm_results}
\end{figure*}
\subsection{Experimental Setup}
\Name is built on top of Triton~\cite{tillet2019triton} (commit: \texttt{0c7edf}) with about 4K lines of C++ and introduces a compact MLIR dialect that encapsulates \aref operations.
We compare \Name against a set of state-of-the-art GPU programming frameworks and libraries, including the closed-source commercial library cuBLAS~\cite{cublas} (v12.7), the NVIDIA open-source CUDA linear algebra library CUTLASS~\cite{cutlass} (v4.0.0), and the baseline Triton under the same commit.
We also include two academic frameworks: ThunderKittens\cite{spector2025thunderkittens} (\texttt{6c27e2}), a CUDA-based AI kernel library, and TileLang\cite{wang2025tilelang} (\texttt{422fb1}), a TVM~\cite{chen2018tvm}-based DSL. To ensure a fair comparison, all baselines are hand-tuned with a fixed set of tile sizes (combinations of $\{64, 128, 256\}$). In the default Triton implementation, the official software pipelining is enabled, with pipeline depth predetermined to balance overlap between communication and computation. For kernels generated by \Name, the size of the \aref and the depth of the MMA pipeline are selected manually to maximize performance, providing each approach with equivalent opportunity to exploit the underlying hardware.

All experiments are conducted on an NVIDIA H100 SXM5 GPU equipped with 80GB of HBM3e memory, using CUDA 12.7 as the software stack.
Experiments are conducted using 25 warp-up and 1000 measurement runs, and we compute the average execution time from these.
We benchmark several representative workloads. The first is GEMM and its variants, evaluated with both FP16 and FP8 precision. The second is an attention kernel, with various sequence lengths to reflect realistic LLM workloads. These benchmarks represent diverse computational patterns and provide a comprehensive assessment of the effectiveness of automatic warp specialization relative to hand-optimized libraries and DSL-based approaches.

\subsection{Matrix Multiplication}
We evaluate GEMM kernels with matrix dimensions of $M=N=8192$ while sweeping the inner dimension $K$ from $256$ to $16384$, and report the throughput for FP16 and FP8.
Notice these matrix sizes are representative in real LLM workloads~\cite{touvron2023llama,dubey2024llama3,grok2,yang2025qwen3}.
As shown in Fig.~\ref{fig:gemm_results}, across both precisions, \Name achieves the same level of performance as the highly optimized kernel library cuBLAS while outperforming the other general-purpose frameworks on most shapes. \Name reaches up to 79\% hardware utilization in both FP16 and FP8.

In FP16, \Name delivers average speedups of 1.01$\times$, 1.13$\times$, 1.15$\times$, and 1.09$\times$ over cuBLAS, Triton, TileLang, and ThunderKittens, respectively.
The improvements over Triton highlight the advantage of overlapping TMA-driven data movement with WGMMA execution. In contrast, Triton employs an Ampere-style software pipelining scheme for asynchronous copies, which is less effective on Hopper than the hardware-backed warp-specialization strategy in \Name.
We also observe \Name is worse than cuBLAS for small $K$, which is mainly because the overhead of Triton becomes relatively significant compared to the highly optimized CUDA library implementation in cuBLAS.
TileLang and ThunderKittens are extensively tuned for large $K$, leading to stronger performance than \Name when $K\geq 8192$.

In FP8, the benefits of automatic warp specialization are more pronounced, particularly at small $K$.
On average, \Name achieves 1.06$\times$, 1.02$\times$, 1.22$\times$, and 1.24$\times$ speedup over cuBLAS, Triton, TileLang, and ThunderKittens, respectively.
This is because the smaller, faster FP8 tiles make Tensor Core computation so rapid that memory transfer and synchronization become bottlenecks. \Name's warp-specialized pipeline alleviates this by maintaining continuous data flow. The \aref abstraction automatically balances producer and consumer warps and deepens prefetching to fully utilize Tensor Cores, yielding larger relative gains for FP8 than for FP16.
TileLang and ThunderKittens appear less tuned for FP8, resulting in up to 1.59$\times$ and 1.61$\times$ lower throughput at small $K$, respectively.
Overall, these results indicate that \Name's \aref abstraction and automatic warp specialization provide a robust orchestration mechanism that overlaps TMA loads and WGMMA across precisions.
We also observe that the Triton baseline exceeds \Name in certain cases, primarily because Triton has extensively tuned its software pipeline for simple GEMM kernels. \Name's advantages become more pronounced in more complex kernels with deeper communication–computation dependencies, such as FlashAttention in \S~\ref{sub:mha}.

\subsection{GEMM Variants}
\begin{figure}[t]
    \centering
    \includegraphics[width=\linewidth]{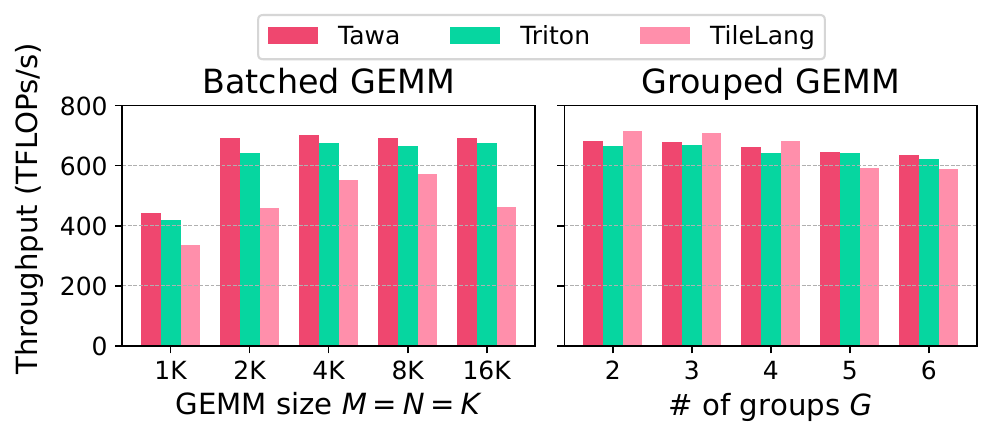}
    \caption{FP16 batched GEMM and grouped GEMM results.}
    \label{fig:grouped_gemm}
\end{figure}
\begin{figure*}[t]
    \centering
    \includegraphics[width=\linewidth]{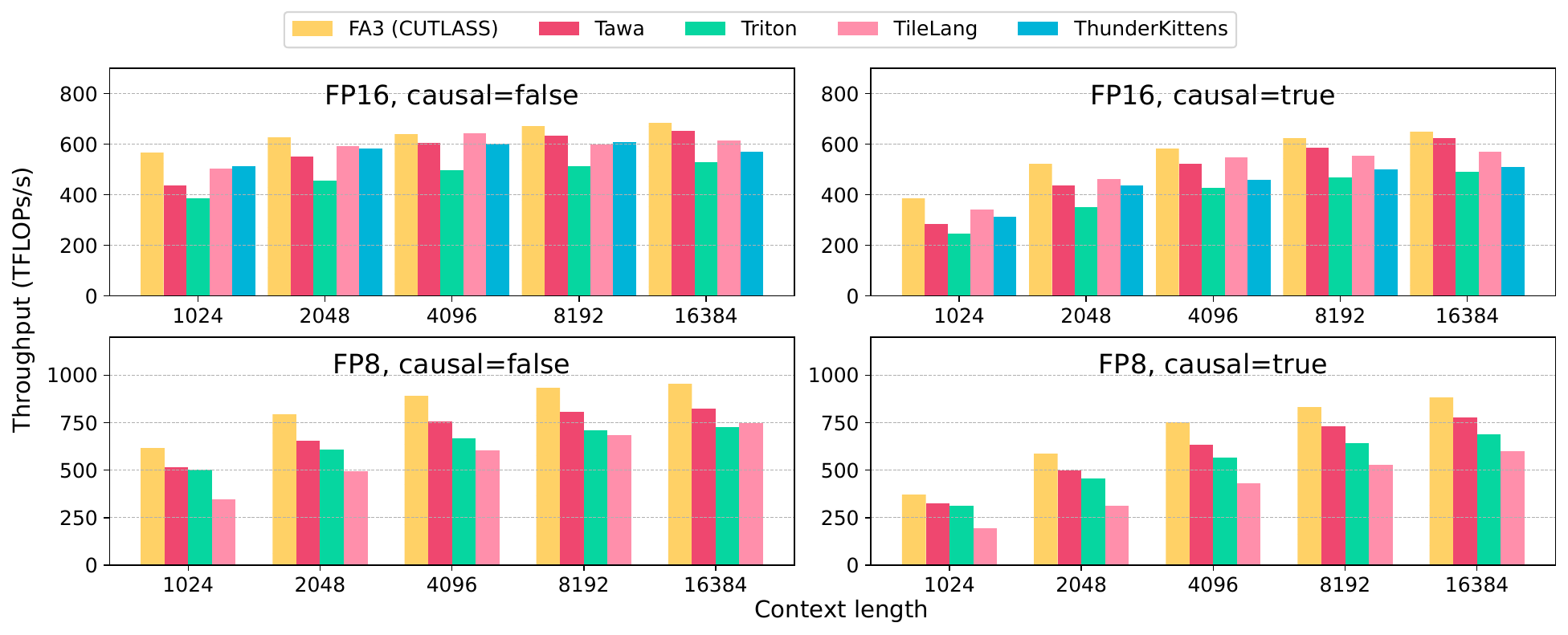}
    \caption{FP16 and FP8 MHA performance results for different frameworks.}
    \label{fig:fmha}
\end{figure*}
To assess the generalizability of \Name across diverse kernel patterns, we evaluate two GEMM variants. The first is batched GEMM, which executes multiple small GEMMs of identical shape within a single kernel. The second is grouped GEMM, which processes multiple GEMMs of varying shapes. Both patterns are pervasive in Mixture of Experts models~\cite{deepseekv3,llama4,grok2}.
We compare against Triton and TileLang; ThunderKittens does not provide functioning kernels for these cases.
For batched GEMM, we fix the batch size to 8 and sweep $M$, $N$, and $K$ from 1K to 16K with square inputs. For grouped GEMM, we vary both the number and shapes of GEMMs, fixing $N$ and $K$ while letting $M$ be a multiple of 512. As shown in Fig.~\ref{fig:grouped_gemm}, \Name consistently outperforms the Triton baseline in both settings, achieving speedups of up to 7\%. Compared to TileLang, \Name is up to 50\% faster on batched GEMM. In grouped GEMM, TileLang performs well for small groups but degrades as group size increases.
These gains stem from \Name's \aref-based partitioning and automatic warp specialization, which allow data movement for one GEMM to be overlapped with computation from another. In batched GEMM, this overlap reduces idle time between consecutive small multiplications, while in grouped GEMM, it schedules heterogeneous shapes more efficiently within a single kernel.

\subsection{Multi-Head Attention}
\label{sub:mha}
We evaluate (non-)causal multi-head attention (MHA) across sequence lengths $L\in[1024, 16384]$ with batch size $4$ and head dimension $128$. As shown in Fig.~\ref{fig:fmha}, \Name is compared against strong reference implementations, including handwritten FlashAttention-3 (FA3)\cite{shah2024fa3} built on \CUTLASS, as well as Triton, TileLang, and ThunderKittens.

\begin{figure}[t]
    \centering
    \includegraphics[width=\linewidth]{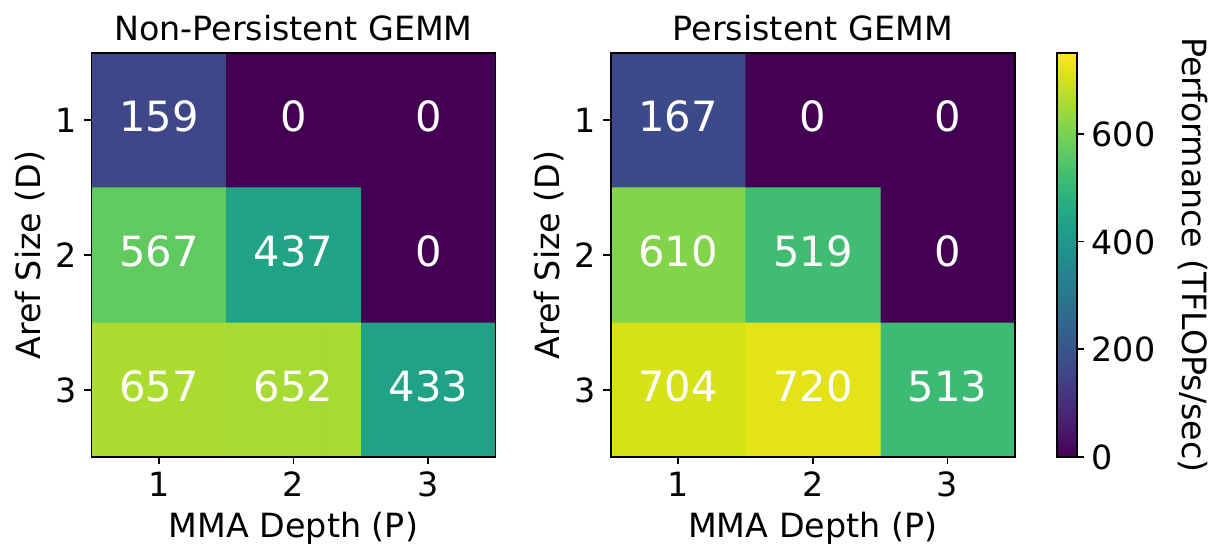}
    \caption{Impact of different \aref and MMA pipeline sizes.}
    \label{fig:hyperparams}
\end{figure}
For FP16, \Name\ attains up to 96\% of FA3 throughput and delivers a consistent 1.21$\times$ speedup over Triton. This gap largely stems from the Triton baseline being effectively a FlashAttention-2 style implementation~\cite{dao2023fa2}, which does not fully exploit Hopper warp specialized producer and consumer pipelines, in particular TMA driven global to shared transfers overlapped with WGMMA on Tensor Cores. At short sequences, the advantage of warp specialization is muted because prologue, epilogue, and barrier costs are not yet amortized. As $L$ grows to at least 4K, memory traffic and synchronization become dominant; in this regime \Name outperforms TileLang and ThunderKittens by up to 1.10$\times$ and 1.23$\times$, respectively, indicating more effective overlap between data movement and compute. Although causal MHA reduces effective Tensor Core utilization due to mask induced control and data hazards, \Name sustains strong performance at long sequences and reaches its peak at $L=16$K.

For FP8, the benefits of automatic warp specialization are even more pronounced at large $L$. \Name reaches up to 89\% of FA3 and delivers 1.11$\times$ and 1.48$\times$ speedup over Triton and TileLang. ThunderKittens fails to run our FP8 attention configurations, indicating that its kernels are primarily tuned for FP16 and lack the layout management and pipeline scheduling needed for efficient FP8 execution.

Overall, the results support two conclusions. First, \aref provides a principled abstraction that converts generic kernels without asynchronous semantics into dataflow pipelines: by treating tiles as references with explicit \aget and \aput phases, \Name overlaps TMA transfers, shared-memory materialization, and WGMMA while maintaining warp-group occupancy and steady MMA throughput. Second, these gains hold across precision from FP16 to FP8 and semantics from noncausal to causal, showing that \Name automatic warp specialization policies generalize well.

\subsection{Hyperparameter Selection}
\label{sub:hyperparam}
To study the impact of hyperparameters in our compilation flow, we conduct experiments on the \aref size $D$, the number of disjoint shared-memory staging buffers governed by asynchronous barriers, and the MMA depth $P$, how many compute tiles (WGMMA issue groups) are overlapped in flight, with (non-)persistent GEMM kernels ($K=16384$).
As shown in Fig.~\ref{fig:hyperparams}, across the feasible region ($D\geq P$), performance consistently increases when increasing $D$. A larger $D$ lets TMA producers prefetch more, smoothing latency variation and better hiding global-to-shared transfer under compute. Further increasing $D$ requires more shared memory to be allocated, which is not achievable for large tile size.

Varying $P$ at fixed $D$ shows the classic over-pipelining trade-off. Increasing $P$ from $1$ to $2$ generally helps (persistent peaks at $D=3,P=2$), because MMA reduces stalls on accumulator availability and amortizes instruction-level latencies. Pushing to $P=3$ lowers throughput: deeper compute pipelines increase live fragments and accumulator registers and can depress WG occupancy, so the marginal latency hiding is outweighed by pressure on registers and shared memory.

Comparing the two panels, the persistent version is consistently faster (roughly +5-10\% on most feasible points). Persistence keeps CTAs resident to pull multiple tiles from a global work queue, which eliminates grid-scheduling and tail effects, improves cache locality for weights and epilogue metadata, and maintains a steady-state TMA and MMA rhythm with fewer disruptions.
Taken together, the figure suggests using a larger $D$ and moderate $P$ ($1$-$2$) to balance overlap and resource pressure, and uses persistent kernels to stabilize the pipeline and harvest cache/reuse benefits.


\begin{figure}[t]
    \centering
    \includegraphics[width=\linewidth]{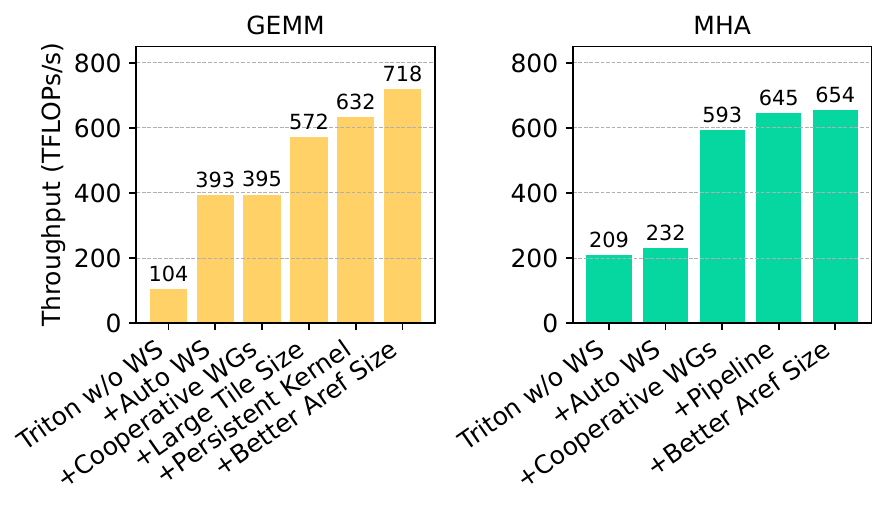}
    \caption{Impact of different optimizations on FP16 kernels.}
    \label{fig:ablation}
\end{figure}
\subsection{Ablation Study}
To further study the effect of different optimization techniques in \Name, we conduct an ablation study on both GEMM and MHA kernels.
We use the largest GEMM ($K=16384$) and MHA kernels ($L=16384$) in our previous experiments, and the results are shown in Fig.~\ref{fig:ablation}.
The baseline Triton without adding warp specialization (WS) delivers a throughput of only 104 TFLOPs/s. By enabling WS with a single warp group while maintaining the same tiling configuration (+Auto WS), performance improves 3.78$\times$ to 393 TFLOPs/s. This substantial jump reflects the efficiency gains from decoupling producer and consumer roles at the warp level, which improves parallelism and mitigates pipeline stalls.

The benefits of warp specialization become even more evident as the configuration is extended to two compute warp groups. With the same tile size of 128$\times$128$\times$64, throughput does not change a lot.
However, the increased number of warps reduces the register pressure and can enable a larger tile size for the warps to cooperatively work on.
When enlarging the tile size to 128$\times$256$\times$64 (+Large Tile Size), it boosts the performance by 1.46$\times$.
The enlarged tile shape improves data reuse and reduces redundant memory traffic, showing that tiling interacts synergistically with warp specialization to exploit the hardware more effectively.
Similar trends can be found in MHA, where the automatic warp specialization (+Auto WS) and cooperative warp groups (+Cooperative WGs) together boost the performance by 2.84$\times$.

Beyond warp specialization and tiling, the introduction of persistent kernels further amplifies performance. In this mode, thread blocks are retained on SMs across multiple iterations, thereby avoiding costly kernel launch overheads and improving data locality. With persistence enabled (+Persistent Kernel), the system further improves the performance by 10\%. Finally, fine-tuning the scheduling depth parameter in conjunction with persistence leads to a peak throughput of 718 TFLOPs/s (+Better Aref Size), representing nearly a seven-fold improvement over the baseline.
For MHA, adding a coarse-grained pipeline and fine-tuning the \aref size leads to the final performance of 654 TFLOPs/s.
This progression highlights how \Name carefully incorporates warp specialization, cooperative warp groups, and pipelining, which can unlock better utilization of modern GPU hardware.

%% file: sections/6-discussion.tex
\section{Future Work}
\label{sec:discussion}
One natural extension of \Name is to broaden the set of communication patterns supported by the compiler. Currently, \Name primarily targets producer-consumer pipelines built around double-buffering, but more advanced schemes such as ping-pong kernels and multicast communication could further improve efficiency. Ping-pong kernels allow alternating roles between warp groups across iterations, thereby balancing workload and alleviating bottlenecks in cases where computation and data movement demands shift dynamically. Similarly, multicasting enables a single producer warp to distribute data to multiple consumer warps simultaneously, reducing redundant memory transfers and better utilizing shared buffers.

Looking further ahead, we aim to generalize \Name beyond Hopper to support next-generation architectures such as Blackwell. In particular, Blackwell introduces tensor memory (\texttt{tmem}) as a new hardware-managed memory hierarchy that complements TMA by providing more flexible data orchestration between shared and global memory.
Leveraging \texttt{tmem} effectively will require extending \aref to model multi-level data movement and revisiting scheduling decisions across heterogeneous memory resources. In addition, future workloads demand even greater scalability, motivating support for multiple producers and consumers for multiple warp groups with distinct roles, as well as more sophisticated graph partitioning algorithms that account for load balancing, register pressure, and memory footprint simultaneously. Specifically, these fine-grained roles can be annotated during the task-aware partitioning process based on instruction semantics as discussed in \S~\ref{sub:partition}.
For example, designating specialized functions such as softmax as special-function warps, while computations outside the main loop form an epilogue warp group that utilizes \texttt{tmem}.

%% file: sections/7-conclusion.tex
\section{Conclusion}
\label{sec:conclusion}
In this paper, we introduced \Name, an automated compiler that generates efficient warp-specialized kernels for modern GPUs using asynchronous references. Our evaluation shows that \Name delivers performance on par with highly optimized handwritten libraries while greatly reducing programming effort. We believe this work offers a step toward principled compiler support for asynchronous GPU programming and can inspire future research on automatically generating high-performance kernels for next-generation architectures.

%% file: sections/ack.tex
\section*{Acknowledgement}
We would like to thank Zihao Ye and the OpenAI Triton team for valuable feedback during the initial design of \Name.
We also thank Rohan Yadav, Matthias Springer, the anonymous reviewers, and our shepherd for their insightful comments on earlier drafts of this manuscript.
Finally, we thank Lei Wang for providing guidance on setting up the TileLang environment.
Hongzheng Chen and Zhiru Zhang are supported in part by ACE, one of the seven centers in JUMP 2.0, a Semiconductor Research Corporation (SRC) program sponsored by DARPA and NSF Award \#2118709.

%% file: main.bbl
\begin{thebibliography}{10}
\providecommand{\url}[1]{#1}
\csname url@samestyle\endcsname
\providecommand{\newblock}{\relax}
\providecommand{\bibinfo}[2]{#2}
\providecommand{\BIBentrySTDinterwordspacing}{\spaceskip=0pt\relax}
\providecommand{\BIBentryALTinterwordstretchfactor}{4}
\providecommand{\BIBentryALTinterwordspacing}{\spaceskip=\fontdimen2\font plus
\BIBentryALTinterwordstretchfactor\fontdimen3\font minus \fontdimen4\font\relax}
\providecommand{\BIBforeignlanguage}[2]{{%
\expandafter\ifx\csname l@#1\endcsname\relax
\typeout{** WARNING: IEEEtran.bst: No hyphenation pattern has been}%
\typeout{** loaded for the language `#1'. Using the pattern for}%
\typeout{** the default language instead.}%
\else
\language=\csname l@#1\endcsname
\fi
#2}}
\providecommand{\BIBdecl}{\relax}
\BIBdecl

\bibitem{amir2024memorywall}
A.~Gholami, Z.~Yao, S.~Kim, C.~Hooper, M.~W. Mahoney, and K.~Keutzer, ``{AI} and memory wall,'' \emph{IEEE Micro}, vol.~44, no.~3, pp. 33--39, 2024.

\bibitem{chen2024understanding}
H.~Chen, J.~Zhang, Y.~Du, S.~Xiang, Z.~Yue, N.~Zhang, Y.~Cai, and Z.~Zhang, ``Understanding the potential of {FPGA}-based spatial acceleration for large language model inference,'' \emph{ACM Transactions on Reconfigurable Technology and Systems}, vol.~18, no.~1, pp. 1--29, 2024.

\bibitem{pope2023scalellmtpu}
R.~Pope, S.~Douglas, A.~Chowdhery, J.~Devlin, J.~Bradbury, J.~Heek, K.~Xiao, S.~Agrawal, and J.~Dean, ``Efficiently scaling transformer inference,'' \emph{Proceedings of machine learning and systems}, vol.~5, pp. 606--624, 2023.

\bibitem{ivanov2021datamovement}
A.~Ivanov, N.~Dryden, T.~Ben-Nun, S.~Li, and T.~Hoefler, ``Data movement is all you need: A case study on optimizing transformers,'' \emph{Proceedings of Machine Learning and Systems}, vol.~3, pp. 711--732, 2021.

\bibitem{chen2024slapo}
H.~Chen, C.~H. Yu, S.~Zheng, Z.~Zhang, Z.~Zhang, and Y.~Wang, ``Slapo: A schedule language for progressive optimization of large deep learning model training,'' in \emph{Proceedings of the 29th ACM International Conference on Architectural Support for Programming Languages and Operating Systems, Volume 2}, ser. ASPLOS'24.\hskip 1em plus 0.5em minus 0.4em\relax New York, NY, USA: Association for Computing Machinery, 2024, pp. 1095--1111.

\bibitem{hopper}
NVIDIA, ``Nvidia hopper architecture,'' \url{https://www.nvidia.com/en-us/data-center/technologies/hopper-architecture/}, 2022.

\bibitem{blackwell}
{NVIDIA}, ``Nvidia blackwell architecture technical brief,'' \url{https://resources.nvidia.com/en-us-blackwell-architecture}, 2025.

\bibitem{tillet2019triton}
P.~Tillet, H.~T. Kung, and D.~Cox, ``Triton: An intermediate language and compiler for tiled neural network computations,'' in \emph{Proceedings of the 3rd ACM SIGPLAN International Workshop on Machine Learning and Programming Languages}.\hskip 1em plus 0.5em minus 0.4em\relax New York, NY, USA: ACM, 2019, pp. 10--19.

\bibitem{cutlass}
NVIDIA, ``Cutlass,'' \url{https://github.com/NVIDIA/cutlass}, 2025.

\bibitem{spector2025thunderkittens}
B.~F. Spector, S.~Arora, A.~Singhal, A.~Parthasarathy, D.~Y. Fu, and C.~Re, ``Thunderkittens: Simple, fast, and adorable kernels,'' in \emph{The Thirteenth International Conference on Learning Representations}, 2025.

\bibitem{shah2024fa3}
J.~Shah, G.~Bikshandi, Y.~Zhang, V.~Thakkar, P.~Ramani, and T.~Dao, ``Flashattention-3: Fast and accurate attention with asynchrony and low-precision,'' \emph{Advances in Neural Information Processing Systems}, vol.~37, pp. 68\,658--68\,685, 2024.

\bibitem{ye2025flashinfer}
Z.~Ye, L.~Chen, R.~Lai, W.~Lin, Y.~Zhang, S.~Wang, T.~Chen, B.~Kasikci, V.~Grover, A.~Krishnamurthy \emph{et~al.}, ``Flashinfer: Efficient and customizable attention engine for {LLM} inference serving,'' \emph{arXiv preprint arXiv:2501.01005}, 2025.

\bibitem{nickolls2008cuda}
J.~Nickolls, I.~Buck, M.~Garland, and K.~Skadron, ``Scalable parallel programming with cuda: Is cuda the parallel programming model that application developers have been waiting for?'' \emph{Queue}, vol.~6, no.~2, pp. 40--53, 2008.

\bibitem{chen2018tvm}
T.~Chen, T.~Moreau, Z.~Jiang, L.~Zheng, E.~Yan, M.~Cowan, H.~Shen, L.~Wang, Y.~Hu, L.~Ceze, C.~Guestrin, and A.~Krishnamurthy, ``Tvm: An automated end-to-end optimizing compiler for deep learning,'' in \emph{Proceedings of the 13th USENIX Conference on Operating Systems Design and Implementation}, ser. OSDI'18.\hskip 1em plus 0.5em minus 0.4em\relax USA: USENIX Association, 2018, p. 579–594.

\bibitem{zheng2020ansor}
L.~Zheng, C.~Jia, M.~Sun, Z.~Wu, C.~H. Yu, A.~Haj-Ali, Y.~Wang, J.~Yang, D.~Zhuo, K.~Sen, J.~E. Gonzalez, and I.~Stoica, ``Ansor: Generating high-performance tensor programs for deep learning,'' in \emph{Proceedings of the 14th USENIX Conference on Operating Systems Design and Implementation}, ser. OSDI'20.\hskip 1em plus 0.5em minus 0.4em\relax USA: USENIX Association, 2020.

\bibitem{shao2022metaschedule}
J.~Shao, X.~Zhou, S.~Feng, B.~Hou, R.~Lai, H.~Jin, W.~Lin, M.~Masuda, C.~H. Yu, and T.~Chen, ``Tensor program optimization with probabilistic programs,'' in \emph{Advances in Neural Information Processing Systems}, 2022.

\bibitem{lai2025relax}
R.~Lai, J.~Shao, S.~Feng, S.~Lyubomirsky, B.~Hou, W.~Lin, Z.~Ye, H.~Jin, Y.~Jin, J.~Liu \emph{et~al.}, ``Relax: composable abstractions for end-to-end dynamic machine learning,'' in \emph{Proceedings of the 30th ACM International Conference on Architectural Support for Programming Languages and Operating Systems, Volume 2}, 2025, pp. 998--1013.

\bibitem{hagedorn2020fireiron}
\BIBentryALTinterwordspacing
B.~Hagedorn, A.~S. Elliott, H.~Barthels, R.~Bodik, and V.~Grover, ``Fireiron: A data-movement-aware scheduling language for {GPUs},'' in \emph{Proceedings of the ACM International Conference on Parallel Architectures and Compilation Techniques}, ser. PACT'20.\hskip 1em plus 0.5em minus 0.4em\relax New York, NY, USA: Association for Computing Machinery, 2020, p. 71–82. [Online]. Available: \url{https://doi.org/10.1145/3410463.3414632}
\BIBentrySTDinterwordspacing

\bibitem{hagedorn2023graphene}
B.~Hagedorn, B.~Fan, H.~Chen, C.~Cecka, M.~Garland, and V.~Grover, ``Graphene: An {IR} for optimized tensor computations on {GPUs},'' in \emph{Proceedings of the 28th ACM International Conference on Architectural Support for Programming Languages and Operating Systems, Volume 3}.\hskip 1em plus 0.5em minus 0.4em\relax New York, NY, USA: Association for Computing Machinery, 2023, pp. 302--313.

\bibitem{wapman2023harmoniccuda}
J.~D. Wapman, S.~Treichler, S.~D. Porumbescu, and J.~D. Owens, ``Harmonic {CUDA}: Asynchronous programming on {GPUs},'' in \emph{Proceedings of the 14th International Workshop on Programming Models and Applications for Multicores and Manycores}, 2023, pp. 39--49.

\bibitem{bauer2011cudadma}
M.~Bauer, H.~Cook, and B.~Khailany, ``C{udaDMA}: optimizing {GPU} memory bandwidth via warp specialization,'' in \emph{Proceedings of 2011 international conference for high performance computing, networking, storage and analysis}, 2011, pp. 1--11.

\bibitem{bauer2014singe}
M.~Bauer, S.~Treichler, and A.~Aiken, ``Singe: Leveraging warp specialization for high performance on {GPUs},'' in \emph{Proceedings of the 19th ACM SIGPLAN symposium on Principles and practice of parallel programming}, 2014, pp. 119--130.

\bibitem{crago2024wasp}
N.~C. Crago, S.~Damani, K.~Sankaralingam, and S.~W. Keckler, ``Wasp: Exploiting gpu pipeline parallelism with hardware-accelerated automatic warp specialization,'' in \emph{2024 IEEE International Symposium on High-Performance Computer Architecture (HPCA)}.\hskip 1em plus 0.5em minus 0.4em\relax IEEE, 2024, pp. 1--16.

\bibitem{chris2021mlir}
\BIBentryALTinterwordspacing
C.~Lattner, M.~Amini, U.~Bondhugula, A.~Cohen, A.~Davis, J.~Pienaar, R.~Riddle, T.~Shpeisman, N.~Vasilache, and O.~Zinenko, ``{MLIR}: Scaling compiler infrastructure for domain specific computation,'' in \emph{Proceedings of the 2021 IEEE/ACM International Symposium on Code Generation and Optimization}, ser. CGO '21.\hskip 1em plus 0.5em minus 0.4em\relax Los Alamitos, CA, USA: IEEE Press, 2021, p. 2–14. [Online]. Available: \url{https://doi.org/10.1109/CGO51591.2021.9370308}
\BIBentrySTDinterwordspacing

\bibitem{yadav2025cypress}
R.~Yadav, M.~Garland, A.~Aiken, and M.~Bauer, ``Task-based tensor computations on modern {GPUs},'' \emph{Proceedings of the ACM on Programming Languages}, vol.~9, no. PLDI, pp. 396--420, 2025.

\bibitem{gluon}
{OpenAI Triton Team}, ``Gluon,'' \url{https://github.com/triton-lang/triton/tree/main/python/examples/gluon}, 2025.

\bibitem{wang2025tilelang}
L.~Wang, Y.~Cheng, Y.~Shi, Z.~Tang, Z.~Mo, W.~Xie, L.~Ma, Y.~Xia, J.~Xue, F.~Yang \emph{et~al.}, ``Tilelang: A composable tiled programming model for ai systems,'' \emph{arXiv preprint arXiv:2504.17577}, 2025.

\bibitem{cheng2025pipethreader}
Y.~Cheng, L.~Wang, Y.~Shi, Y.~Xia, L.~Ma, J.~Xue, Y.~Wang, Z.~Mo, F.~Chen, F.~Yang, M.~Yang, and Z.~Yang, ``Pipethreader: software-defined pipelining for efficient dnn execution,'' in \emph{Proceedings of the 19th USENIX Conference on Operating Systems Design and Implementation}, ser. OSDI '25.\hskip 1em plus 0.5em minus 0.4em\relax USA: USENIX Association, 2025.

\bibitem{paszke2019pytorch}
A.~Paszke, S.~Gross, F.~Massa, A.~Lerer, J.~Bradbury, G.~Chanan \emph{et~al.}, ``Pytorch: An imperative style, high-performance deep learning library,'' in \emph{Proceedings of the 33rd International Conference on Neural Information Processing Systems}.\hskip 1em plus 0.5em minus 0.4em\relax New York, NY, USA: IEEE Press, 2019, pp. 172--198.

\bibitem{pytorch20}
J.~Ansel, E.~Yang, H.~He, N.~Gimelshein, A.~Jain, M.~Voznesensky \emph{et~al.}, ``Pytorch 2: {Faster Machine Learning Through Dynamic Python Bytecode Transformation and Graph Compilation},'' in \emph{Proceedings of the 29th ACM International Conference on Architectural Support for Programming Languages and Operating Systems (ASPLOS'24)}.\hskip 1em plus 0.5em minus 0.4em\relax New York, NY, USA: Association for Computing Machinery, 2024, pp. 317--335.

\bibitem{bastian2025asyncgraphene}
B.~Hagedorn and V.~Grover, ``It's about time - temporal abstractions for asynchronous gpu tensor computation,'' In submission, 2025.

\bibitem{opencl}
K.~Groups, ``{OpenCL} 2.0 reference guide,'' \url{https://www.khronos.org/assets/uploads/developers/presentations/opencl20-quick-reference-card.pdf}, 2015.

\bibitem{oneapi}
Intel, ``{oneAPI} toolkits: Pipe,'' \url{https://www.intel.com/content/www/us/en/docs/oneapi-fpga-add-on/optimization-guide/2023-1/the-pipe-class-and-its-use.html}, 2023.

\bibitem{vitis}
AMD, ``Vitis hls stream library,'' \url{https://docs.amd.com/r/en-US/ug1399-vitis-hls/HLS-Stream-Library}, 2025.

\bibitem{fang2025dato}
S.~Fang, H.~Chen, N.~Zhang, J.~Li, H.~Meng, A.~Liu, and Z.~Zhang, ``Dato: A task-based programming model for dataflow accelerators,'' \emph{arXiv preprint arXiv:2509.06794}, 2025.

\bibitem{chen2024allo}
H.~Chen, N.~Zhang, S.~Xiang, Z.~Zeng, M.~Dai, and Z.~Zhang, ``Allo: A programming model for composable accelerator design,'' \emph{Proceedings of the ACM on Programming Languages}, vol.~8, no. PLDI, pp. 593--620, 2024.

\bibitem{xiang2022heteroflow}
\BIBentryALTinterwordspacing
S.~Xiang, Y.-H. Lai, Y.~Zhou, H.~Chen, N.~Zhang, D.~Pal, and Z.~Zhang, ``Heteroflow: An accelerator programming model with decoupled data placement for software-defined fpgas,'' in \emph{Proceedings of the 2022 ACM/SIGDA International Symposium on Field-Programmable Gate Arrays}, ser. FPGA'22.\hskip 1em plus 0.5em minus 0.4em\relax New York, NY, USA: Association for Computing Machinery, 2022, p. 78–88. [Online]. Available: \url{https://doi.org/10.1145/3490422.3502369}
\BIBentrySTDinterwordspacing

\bibitem{vaswani2017transformer}
A.~Vaswani, N.~Shazeer, N.~Parmar, J.~Uszkoreit, L.~Jones, A.~N. Gomez, L.~Kaiser, and I.~Polosukhin, ``Attention is all you need,'' in \emph{Proceedings of the 31st International Conference on Neural Information Processing Systems}, ser. NIPS'17.\hskip 1em plus 0.5em minus 0.4em\relax Red Hook, NY, USA: Curran Associates Inc., 2017, p. 6000–6010.

\bibitem{cublas}
NVIDIA, ``{cuBLAS}: Basic linear algebra on {NVIDIA GPUs},'' \url{https://developer.nvidia.com/cublas}, 2025.

\bibitem{touvron2023llama}
H.~Touvron, T.~Lavril, G.~Izacard, X.~Martinet, M.-A. Lachaux, T.~Lacroix, B.~Rozi{\`e}re, N.~Goyal, E.~Hambro, F.~Azhar \emph{et~al.}, ``Llama: Open and efficient foundation language models,'' \emph{arXiv preprint arXiv.2302.13971}, 2023.

\bibitem{dubey2024llama3}
\BIBentryALTinterwordspacing
A.~Dubey, A.~Jauhri, A.~Pandey, A.~Kadian, A.~Al-Dahle, A.~Letman, A.~Mathur, A.~Schelten, A.~Yang, A.~Fan \emph{et~al.}, ``The llama 3 herd of models,'' 2024. [Online]. Available: \url{https://arxiv.org/abs/2407.21783}
\BIBentrySTDinterwordspacing

\bibitem{grok2}
xAI, ``Grok 2,'' \url{https://x.ai/news/grok-2}, 2025.

\bibitem{yang2025qwen3}
\BIBentryALTinterwordspacing
A.~Yang, A.~Li, B.~Yang, B.~Zhang, B.~Hui, B.~Zheng, B.~Yu, C.~Gao, C.~Huang, C.~Lv, C.~Zheng, D.~Liu, F.~Zhou, F.~Huang, F.~Hu, H.~Ge, H.~Wei, H.~Lin, J.~Tang, J.~Yang, J.~Tu, J.~Zhang, J.~Yang, J.~Yang, J.~Zhou, J.~Zhou, J.~Lin, K.~Dang, K.~Bao, K.~Yang, L.~Yu, L.~Deng, M.~Li, M.~Xue, M.~Li, P.~Zhang, P.~Wang, Q.~Zhu, R.~Men, R.~Gao, S.~Liu, S.~Luo, T.~Li, T.~Tang, W.~Yin, X.~Ren, X.~Wang, X.~Zhang, X.~Ren, Y.~Fan, Y.~Su, Y.~Zhang, Y.~Zhang, Y.~Wan, Y.~Liu, Z.~Wang, Z.~Cui, Z.~Zhang, Z.~Zhou, and Z.~Qiu, ``Qwen3 technical report,'' 2025. [Online]. Available: \url{https://arxiv.org/abs/2505.09388}
\BIBentrySTDinterwordspacing

\bibitem{deepseekv3}
\BIBentryALTinterwordspacing
DeepSeek-AI \emph{et~al.}, ``Deepseek-v3 technical report,'' 2025. [Online]. Available: \url{https://arxiv.org/abs/2412.19437}
\BIBentrySTDinterwordspacing

\bibitem{llama4}
Meta, ``The {Llama 4} herd: The beginning of a new era of natively multimodal ai innovation,'' \url{https://ai.meta.com/blog/llama-4-multimodal-intelligence/}, 2025.

\bibitem{dao2023fa2}
T.~Dao, ``Flashattention-2: Faster attention with better parallelism and work partitioning,'' \emph{arXiv preprint arXiv:2307.08691}, 2023.

\end{thebibliography}
